\pdfoutput=1

\documentclass[11pt]{article}

\usepackage{acl}

\usepackage{times}
\usepackage{latexsym}
\usepackage{graphicx}
\usepackage{tikz}
\usepackage{amsmath}
\usepackage{booktabs}
\usepackage{enumitem}
\usepackage{array,multirow, multicol, colortbl, subcaption, makecell, tabularx, amsfonts}

\usepackage{arydshln}

\newcommand{\minisection}[1]{\noindent{\bf #1}} 
\usepackage{xurl}

\usepackage[T1]{fontenc}

\usepackage[utf8]{inputenc}

\usepackage{microtype}

\title{Representation of Lexical Stylistic Features \\ in Language Models' Embedding Space}

\author{Qing Lyu \hspace{7mm} Marianna Apidianaki \hspace{7mm} Chris Callison-Burch \\
    University of Pennsylvania \\
    \texttt{\{lyuqing, marapi, ccb\}@seas.upenn.edu}
    }

\begin{document}
\maketitle
\begin{abstract}

The representation space of pretrained Language Models (LMs) encodes rich information about words and their relationships (e.g., similarity, hypernymy, polysemy) as well as abstract semantic notions (e.g., intensity). In this paper, we demonstrate that lexical stylistic notions such as complexity, formality, and figurativeness, can also be identified in this space. We show that it is possible to derive a vector representation for each of these stylistic notions from only a small number of seed pairs. Using these vectors, we can characterize new texts in terms of these dimensions by performing simple calculations in the corresponding embedding space. We conduct experiments on five datasets and find that static embeddings encode these features more accurately at the level of words and phrases, whereas contextualized LMs perform better on sentences. The lower performance of contextualized representations at the word level is partially attributable to the anisotropy of their vector space, which can be corrected to some extent using techniques like standardization.\footnote{Our code and data are publicly available at \url{https://github.com/veronica320/Lexical-Stylistic-Features}.}

\end{abstract}

\section{Introduction}

The style of a text is often reflected in its grammatical and discourse properties, but also in local word choices made by the author. The choice of one from a set of synonyms or paraphrases with different connotations can define the style of a text in terms of complexity (e.g., \textit{help} vs. \textit{assist}), formality (e.g., \textit{dad} vs. \textit{father}), figurativeness (e.g., \textit{fall} vs. \textit{plummet}), and so on \citep{10.1162/089120102760173625}.
These \textit{lexical stylistic features} can be useful in various scenarios, such as analyzing the style of authors or texts of different genres, and determining the appropriate word usage in language learning applications. 

\begin{table}[t]
\centering
    \scalebox{0.8}{
    \begin{tabular}{|c|c|}
    \toprule
     \parbox[t]{2mm}{\centering \multirow{7}{*}{\rotatebox[origin=c]{90}{{\sc Complexity}}}}
    & doctor $\rightarrow$ medical practitioner \\ 
    & laws $\rightarrow$ legislative texts \\ 
    & high blood pressure $\rightarrow$ hypertension \\ 
    & very common $\rightarrow$ prevalent \\ 
    & a lot $\rightarrow$ significant quantity \\ 
    & be bad $\rightarrow$ impact negatively \\
    & help $\rightarrow$ assist \\ 
\hline
 \parbox[t]{2mm}{\centering \multirow{7}{*}{\rotatebox[origin=c]{90}{{\sc Formality}}}}
    & my gosh $\rightarrow$ jesus \\ 
    & breathing $\rightarrow$ respiratory \\ 
    & yeah $\rightarrow$ yes \\ 
    & ten years $\rightarrow$ decade \\ 
    & first of all $\rightarrow$ foremost \\ 
    & a whole bunch $\rightarrow$ full \\ 
    & my dad $\rightarrow$ father \\ \hline
 \parbox[t]{2mm}{\centering \multirow{7}{*}{\rotatebox[origin=c]{90}{{\sc Figurativeness}}}}
    & bright $\rightarrow$ radiant \\ 
    & heavy $\rightarrow$ burdened \\ 
    & unsympathetic $\rightarrow$ cold-hearted	 \\ 
    & fall $\rightarrow$ plummet \\ 
    & a lot of $\rightarrow$ a sea of	 \\ 
    & quick $\rightarrow$ lightning \\ 
    & hard $\rightarrow$ ironclad \\ 
     \bottomrule
    \end{tabular}}
    \caption{Seed pairs for constructing vector representations of complexity (simple $\rightarrow$ complex), formality (casual $\rightarrow$ formal), and figurativeness (literal $\rightarrow$ figurative).
    }
    \vspace{-0.15in}
    \label{table:seedpairs}
\end{table}

Previous approaches to formality detection relied on word length, frequency, as well as on the presence of specific prefixes and suffixes (e.g., {\it intra-}, {\it -ation}) \cite{brooke-etal-2010-automatic}. Such features have also been used for complexity detection, often combined with information regarding the number of word senses and synonyms \cite{shardlow:2013:SRW,kriz-etal-2018-simplification}. Recent studies have shown that the representation space of pretrained LMs encodes a wealth of lexical semantic information, including similarity, polysemy, and hypernymy \cite[][i.a.]{GariSolerandApidianaki:TACL2021,pimentel-etal-2020-speakers,10.1162/tacl_a_00298,ravichander-etal-2020-systematicity,vulic-etal-2020-probing}. In particular, abstract semantic notions such as intensity (e.g., {\it pretty} $\rightarrow$ {\it beautiful} $\rightarrow$ {\it gorgeous}) can be extracted using a lightweight approach based on simple calculations in the vector space \cite{gari-soler-apidianaki-2020-bert,gari-soler-apidianaki-2021-scalar}. 

In this paper, we explore whether lexical stylistic features can also be identified in the vector space built by pretrained LMs. To do this, we extend the method of \citet{gari-soler-apidianaki-2020-bert} to address complexity, formality, and figurativeness. We first construct a vector representation for each of these features using a small number of seed pairs shown in Table~\ref{table:seedpairs}. We then use these vectors to characterize new texts according to these stylistic dimensions, by applying simple calculations in the vector space. We evaluate our method using a binary classification task: given a pair of texts that are semantically similar but stylistically different in terms of some target feature (e.g., formality), the task is to determine which text exhibits the feature more strongly (e.g., is more formal). Note that the goal of our study is not to achieve high performance on the task itself, but rather to probe for how well these stylistic features are encoded in different types of pretrained representations. 

We experiment with various static and contextualized embeddings on five datasets, containing words and phrases ({\it doctor} vs. {\it medical practitioner}), or sentences ({\it Those recommendations were unsolicited and undesirable.} vs. {\it that’s the stupidest suggestion EVER.}). Our results show that both types of representations can capture these stylistic features reasonably well, although static embeddings perform better at the word and phrase level, and contextualized LMs at the sentence level. We hypothesize that the sub-optimal performance of contextualized LMs on short texts might be partially due to the high \textit{anisotropy} of their embedding space. Anisotropic word representations occupy a narrow cone instead of being uniformly distributed in the vector space, resulting in highly positive correlations even for unrelated words, thus negatively impacting the quality of the similarity estimates that can be drawn from the space \cite{ethayarajh-2019-contextual,gao2018representation,cai2021isotropy,rajaee-pilehvar-2021-cluster}. We verify this hypothesis by implementing different anisotropy correction strategies \citep{timkey-van-schijndel-2021-bark} and discuss the observed improvements in contextualized representations' performance on short texts. 

Overall, our findings contribute to the big picture of probing literature, showing that stylistic features like complexity, formality, and figurativeness can be decoded from the embedding space of pretrained representations using simple calculations, without any supervision. Our lightweight method can be easily integrated into downstream applications like authorship attribution and style transfer.

\section{Related work}

There has been an extensive body of literature on probing techniques aimed at identifying the linguistic and world knowledge encoded in LM representations.
For example, given a Machine Translation model, does it implicitly capture the syntax structure of the source text? Existing work addresses such questions with methods like auxiliary classifiers (a.k.a. probing/diagnostic classifiers) \cite{veldhoen2016diagnostic,adi_fine-grained_2017,conneau_what_2018}, information-theoretic probing \cite{voita_information-theoretic_2020,lovering_information-theoretic_2020}, behavioral tests \cite{ebrahimi_hotflip_2018,wallace_universal_2019,petroni_language_2019}, geometric probing \cite{chang-etal-2022-geometry, wartena-2022-geometry, Kozlowskietal:2019}, visualization of model-internal structures \cite{raganato-tiedemann-2018-analysis}, and so on. Using these methods, researchers have found that pretrained LMs do encode various types of knowledge, including syntactic \cite{linzen_assessing_2016,hewitt_structural_2019}, semantic \cite{ettinger_probing_2016,adi_fine-grained_2017,yanaka_neural_2020}, pragmatic \cite{jeretic_are_2020,schuster_harnessing_2020}, as well as factual and commonsense knowledge \cite{petroni_language_2019,thukral_probing_2021}. 

Our work is along the line of probing for lexical semantics with simple geometry-based methods \cite{vulic-etal-2020-probing, GariSolerandApidianaki:TACL2021}, which uncovers the target knowledge encoded in the semantic space of LM representations with simple geometric computations \cite{vulic-etal-2020-probing, GariSolerandApidianaki:TACL2021}. Compared to the most widely used auxiliary classifier method, geometric probing does not rely on any external model. Thus, it requires no annotated training data and avoids the potential issue of the external model itself learning the target knowledge \cite{hewitt-liang-2019-designing}.

Directly related to our work, \citet{gari-soler-apidianaki-2020-bert} proposed a method to detect the intensity of scalar adjectives, where an ``intensity'' dimension is identified in the vector space built by the BERT model. The method draws inspiration from word analogies in gender bias work, where a gender subspace is identified in the embedding space by calculating the main 
direction spanned by the differences between vectors of gendered word pairs (e.g., $\overrightarrow{he}$ - $\overrightarrow{she}$, $\overrightarrow{man}$ - $\overrightarrow{woman}$) 
\cite{Bolukbasi-NIPS2016,devandphilips2019}. Similarly,
\citet{gari-soler-apidianaki-2020-bert} view intensity as a direction in the embedding space which is calculated by subtracting the vector of a low-intensity adjective from that of a high-intensity adjective on the same scale (e.g., $\overrightarrow{awesome}$ - $\overrightarrow{good}$, $\overrightarrow{horrible}$ - $\overrightarrow{bad}$). Intuitively, this subtraction cancels out the adjectives' common denotation and retains their variance in intensity, which is represented by the resulting difference vector ($\overrightarrow{dVec}$).
This vector can then be used to determine the intensity of new adjectives by simply taking the cosine similarity of their vector to $\overrightarrow{dVec}$. We extend this method to other lexical stylistic notions, and address words of different part-of-speech (POS) and longer texts.

\section{Method}
\label{sec:method}

\begin{figure}
  \centering
  \includegraphics[width=0.9\columnwidth]{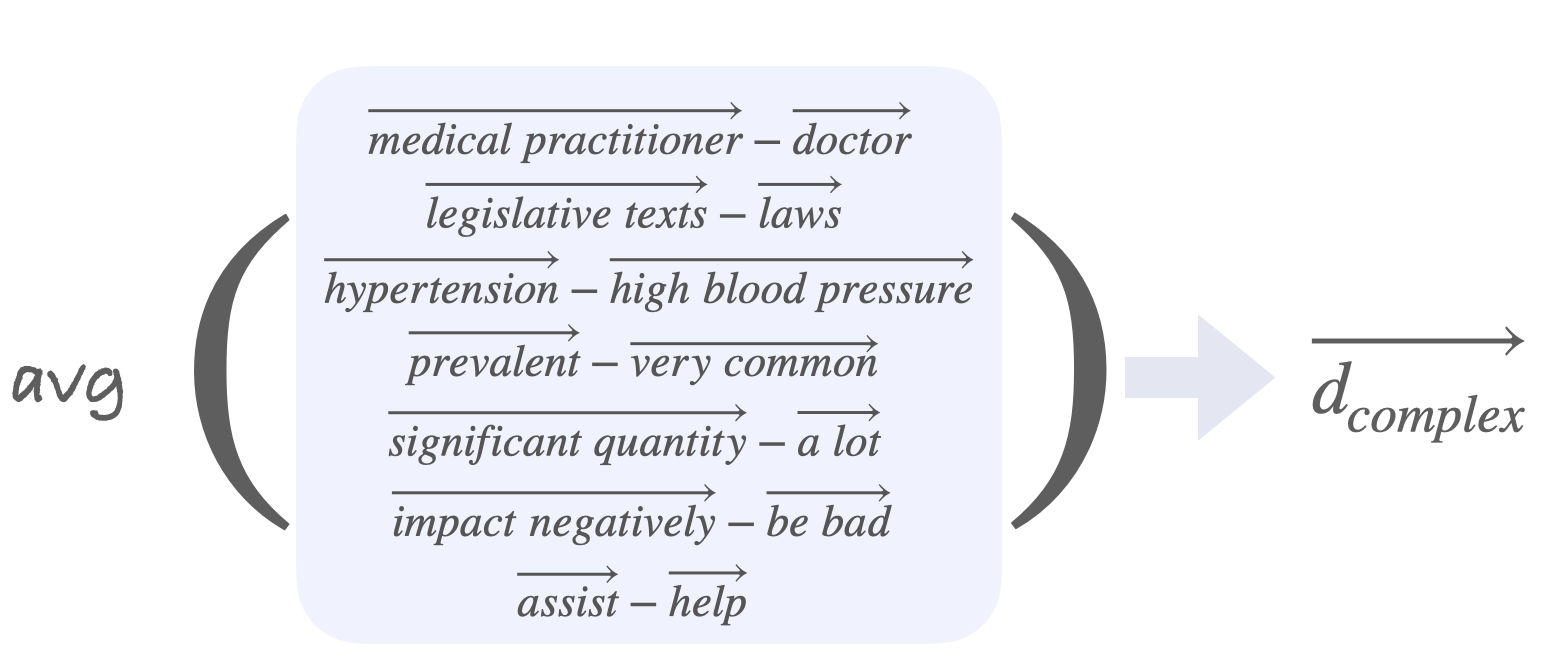}
  \caption{Complexity vector generation.}
  \label{fig:dcomplex}
\end{figure}

We adopt the definitions for the three stylistic features of interest (complexity, formality, and figurativeness) proposed by previous work. Simple language is ``used to talk to children or non-native English speakers'', whereas more complex language is ``used by academics or domain experts'' \cite{pavlick-nenkova-2015-inducing}. Formal language is defined as ``the way one talks to a superior'', whereas casual language is ``used with friends'' \cite{pavlick-nenkova-2015-inducing}. Figurative language is defined by \citet{stowe-etal-2022-impli} as utterances ``in which the intended meaning differs from the literal compositional meaning'', while literal language exhibits no such difference.  
Unlike the previous two features, figurativeness is often a contextual instead of lexical feature (e.g., the word {\it adhere} is used in a metaphorical sense in the expression ``{\it adhere} to the rules'' and in its literal sense in ``{\it adhere} to the wall'').\footnote{In the literature, figurativeness is generally studied at the level of utterances \cite{stowe-etal-2022-impli,piccirilli-schulte-im-walde-2022-drives,chakrabarty-etal-2022-flute}. 
Some studies also look at the semantic properties of words and phrases as indicators for metaphor identification 
\cite{birke-sarkar-2006-clustering,tsvetkov-etal-2013-cross,gutierrez-etal-2016-literal}.} We explore the usability of our method for studying figurativeness by using a small seed set of synonyms and paraphrases that have literal and figurative connotations (e.g., {\it unsympathetic} $\rightarrow$ {\it cold-hearted}) independent of their context. These pairs are only used for constructing our figurativeness vector representation, while our evaluation is performed on a dataset containing full sentences (see Section~\ref{sec:exp_setup} for details). 

Our method involves two steps: (a) \textbf{feature vector generation}, where we construct a vector representation for each feature; and (b) \textbf{feature value prediction}, where we predict how strongly a new piece of text exhibits some target feature using the constructed feature vector. We illustrate the two steps below.

\paragraph{Feature vector generation.} 

We collect a small number of seed pairs to illustrate each notion, shown in Table \ref{table:seedpairs}.\footnote{This is based on the finding from \citet{gari-soler-apidianaki-2020-bert} that using only a few or even a single pair(s) is almost as competitive as using an entire corpus in the case of intensity ranking.} The seed pairs consist of rough paraphrases that differ in the stylistic aspect of interest. Consider complexity as an example. Given a pair of ``simple $\rightarrow$ complex'' texts, we subtract the vector of the simple from that of the complex one (e.g., $\overrightarrow{medical\; practitioner}$ - $\overrightarrow{doctor}$). After performing this subtraction for each pair in the seed set, we then average the resulting difference vectors to obtain a vector representing complexity which we call $\overrightarrow{d_{complex}}$. This procedure is illustrated in Figure \ref{fig:dcomplex}. 
Similarly, for formality, we subtract the vector of the informal paraphrase from that of its formal counterpart (e.g., $\overrightarrow{respiratory}$ - $\overrightarrow{breathing}$), and for figurativeness, we subtract the vector of the literal expression from that with figurative meaning (e.g., $\overrightarrow{bright}$ - $\overrightarrow{radiant}$). 
By averaging the difference vectors for all pairs in the corresponding seed set, we obtain vectors representing formality ($\overrightarrow{d_{formal}}$) and figurativeness ($\overrightarrow{d_{fig}}$). 
We extend the method of \citet{gari-soler-apidianaki-2020-bert}, which was only applied to scalar adjectives, to words of other POS and to longer text (phrases and sentences). Finally, we compare the vectors that are built using representations from different monolingual and multilingual models. 

\paragraph{Feature value prediction.} Given a new piece of text (word, phrase, or sentence), we compute the cosine similarity between its vector representation and $\overrightarrow{d_{complex}}$, $\overrightarrow{d_{formal}}$ and $\overrightarrow{d_{fig}}$. The more similar the vector of the new text is to one of these feature vectors, the more complex, formal, or figurative the text is considered to be.

\section{Experimental Setup}
\label{sec:exp_setup}

\minisection{Evaluation task and metrics.} We evaluate the representation of the target features in a binary classification task: given a pair of texts (words, phrases, or sentences) $t_0$ and $t_1$ that are \textbf{semantically similar} but \textbf{stylistically different} in terms of some feature $F$ (e.g., figurativeness), the task is to decide which text exhibits the feature more strongly (e.g., is more figurative). For example, given two sentences ``You must adhere to the rules.'' ($t_0$) and ``You must obey the rules.'' ($t_1$), the ground truth is that $t_0$ is more figurative. We use accuracy as our evaluation metric.

\vspace{1mm}

\minisection{Seed pairs.} For each feature, we use seven seed pairs for vector generation, as shown in Table~\ref{table:seedpairs}. The seeds for complexity are examples from the paper describing SimplePPDB \cite{pavlick-callison-burch-2016-simple}, and the seeds for formality are from the paper on lexical style properties of paraphrases \cite{pavlick-nenkova-2015-inducing}. For figurativeness, we manually compile a set of seven seed pairs.

\begin{table}[!t]
    \centering
    \scalebox{0.9}{
    \begin{tabular}{p{0.2\columnwidth}>{\centering\arraybackslash}p{0.35\columnwidth}>{\centering\arraybackslash}p{0.35\columnwidth}}
    \toprule
        \textbf{Feature} & \textbf{Short-text} (word/phrase) & \textbf{Long-text} \newline (sentence) \\ \hline
        Complexity & SimplePPDB & SimpleWikipedia \\ 
        Formality & StylePPDB & GYAFC  \\ 
        Figurativeness & – & IMPLI \\
        \bottomrule
    \end{tabular}
    }
    \caption{Datasets used for each feature.}
    \label{table:datasets}
\end{table}

\vspace{1mm}
\minisection{Datasets.} The datasets used in our feature value prediction experiments (described in Table~\ref{table:datasets}) contain pairs of words or phrases (short text), and pairs of sentences (long text). Note that this distinction is not based on the number of tokens, but on whether the text is a complete sentence. For complexity, we use SimplePPDB \cite{pavlick-callison-burch-2016-simple} and SimpleWikipedia \cite{kauchak-2013-improving}; for formality, Style-annotated PPDB (StylePPDB for short) \cite{pavlick-nenkova-2015-inducing} and GYAFC \cite{rao-tetreault-2018-dear}. For figurativeness, since there is no dataset of word and/or phrase pairs, we only use the IMPLI (Idiomatic and Metaphoric Paired Language Inference) dataset \cite{stowe-etal-2022-impli} that contains sentences.

For each dataset, we select the optimal configuration (see the Configuration paragraph below) using the validation set, and report its performance on the test set. 
To make the label distribution balanced, we randomly shuffle the order of the two pieces of text in each pair and re-assign the gold label accordingly. This ensures that a majority baseline only performs around chance. Figure~\ref{fig:freq_stats} shows the distribution of token frequency in each dataset.\footnote{See Appendix~\ref{sec:appendix_dataset} for more details including dataset statistics, pre-processing method, dataset splits, and examples.}

\begin{figure}[t!]
    \centering
    \scalebox{0.9}{
    \includegraphics[width=0.9\columnwidth]{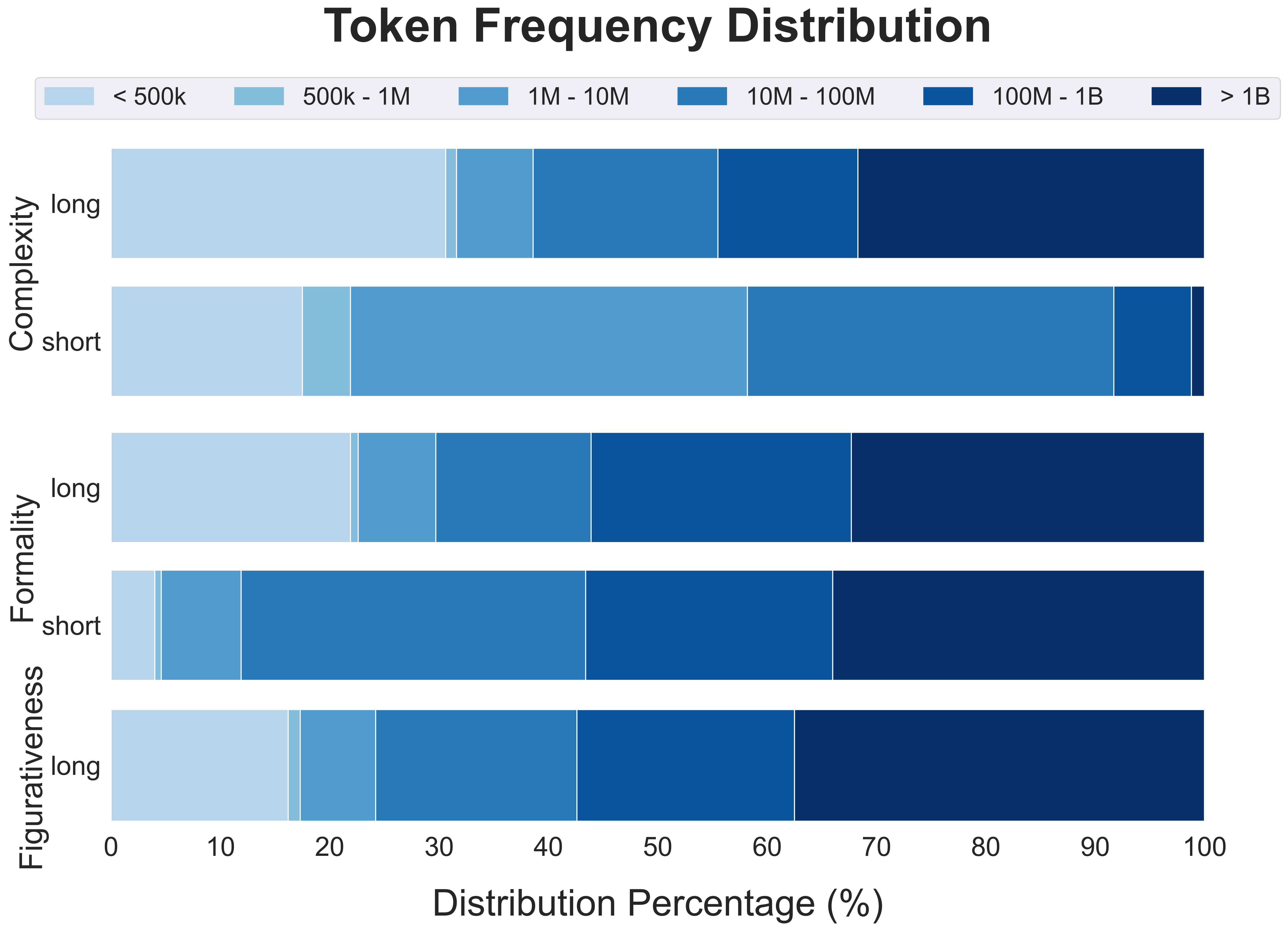}
    }
    \caption{Distribution of token frequency in the evaluation datasets.}
    \label{fig:freq_stats}
\end{figure}

\vspace{1mm}
\minisection{Baselines.} We compare our method to two simple baselines. The \textbf{majority baseline} always predicts the majority label in the dataset. The \textbf{frequency baseline} consults the frequency counts of each token in the Google N-gram corpus \cite{brants2006web} and considers more frequent tokens to be simpler, more casual, and more literal. Frequency has been a strong baseline for complexity and formality in previous work, given that rare words tend to be more complex than frequently used words \cite{brooke-etal-2010-automatic}.

\begin{table*}[!t]
    \centering
    \scalebox{0.7}{

    \begin{tabular}{ll|>{\raggedright\arraybackslash}p{2.5cm}>{\raggedright\arraybackslash}p{2.5cm}|>{\raggedright\arraybackslash}p{2.5cm}>{\raggedright\arraybackslash}p{2.5cm}|>{\raggedright\arraybackslash}p{2.5cm}}

    \toprule
        \textbf{Pooling} & \parbox{4.3cm}{\centering \textbf{Model}} & \multicolumn{2}{c|}{\textbf{Complexity}} & \multicolumn{2}{c|}{\textbf{Formality}} & \textbf{Figurativeness} \\
        ~ & ~ & short & long & short & long & long \\
         \hline
        ~ & majority & 55.1 & 50.6 & 51.2 & 51.8 & 51.4 \\
        \hline
        \multirow{6}{1.5cm}{Mean} & frequency & 83.2 & 51.0 & 61.0 & 41.4 & 49.7 \\ \cdashline{2-7}
        ~ & \multirow{2}{4cm}{static} & 84.8 \newline \footnotesize{glove} & 60.0 \newline \footnotesize{glove} & \textbf{76.8} \newline \footnotesize{glove} & 82.8 \newline \footnotesize{glove} & 54.3 \newline \footnotesize{glove} \\ \cdashline{2-7}
        ~ & \multirow{2}{4.3cm}{contextualized (single layer)} & \textbf{86.2} \newline \footnotesize{roberta-large (4)} & \textbf{76.5} \newline \footnotesize{mbert-base (1)} & 68.7 \newline \footnotesize{bert-base (1)} & 82.4 \newline \footnotesize{roberta-large (12)} & \textbf{72.9} \newline \footnotesize{bert-large (14)} \\ \cdashline{2-7} 
        ~ & \multirow{2}{4cm}{contextualized (layer agg)} & 84.4 \newline \footnotesize{mbert-base (10)} & 76.0 \newline \footnotesize{mbert-base (11)} & 67.6 \newline \footnotesize{bert-large (1)} & \textbf{86.7} \newline \footnotesize{roberta-large (23)} & 67.2 \newline \footnotesize{bert-large (19)} \\
        \hline
        \multirow{6}{1.5cm}{Max} & frequency & 80.7 & 46.4 & 57.2 & 42.5 & 47.9 \\ \cdashline{2-7}
        ~ & \multirow{2}{4cm}{static} & \textbf{89.4} \newline \footnotesize{glove}  & 58.0 \newline \footnotesize{glove} & \textbf{76.0} \newline \footnotesize{glove} & 63.4 \newline \footnotesize{glove}  & 56.0 \newline \footnotesize{fasttext} \\ \cdashline{2-7}
        ~ & \multirow{2}{4.3cm}{contextualized (single layer)} & 87.7 \newline \footnotesize{roberta-large (4)} & \textbf{69.4} \newline \footnotesize{roberta-base (12)}  & 71.7 \newline \footnotesize{mbert-base (0)} & \textbf{73.6} \newline \footnotesize{mbert-base (1)} & \textbf{64.8} \newline \footnotesize{bert-large (11)} \\ \cdashline{2-7}
        ~ & \multirow{2}{4cm}{contextualized (layeragg)} & 86.2 \newline \footnotesize{roberta-large (19)} & 67.6 \newline \footnotesize{roberta-large (4)} & 71.7\newline \footnotesize{mbert-base (0)}  & 71.7 \newline \footnotesize{roberta-large (24)}  & 63.9 \newline \footnotesize{bert-large (14)}  \\ 
        \bottomrule
    \end{tabular}
    }
    \caption{Accuracy scores obtained on each test set using different types of embeddings and pooling methods. We report the performance of the models and layers (in parentheses) that best predicted the feature on the corresponding validation set. For contextualized representations, we report results using a single layer or layer aggregation (``layer agg''). The highest performance obtained with each pooling method (Mean/Max) is in boldface.}
    \label{table:main_results}
\end{table*}
\begin{table}[!t]
    \centering
    \scalebox{0.65}{
    \begin{tabular}{ll|cc|cc|c}
    \toprule
        \textbf{Pooling} & \textbf{Stats} & \multicolumn{2}{c|}{\textbf{Complexity}} & \multicolumn{2}{c|}{\textbf{Formality}} & \textbf{Figurativeness} \\
        ~ & ~ & short & long & short & long & long \\
         \hline
        \multirow{2}{1.4cm}{Mean} & 2 beats 1 (\%) & 63.0 & 78.0 & 92.9 & 72.4 & 54.3 \\ 
        ~ & acc gain & \cellcolor{green!20} 2.6 & \cellcolor{green!20} 4.1 & \cellcolor{green!20} 4.3 & \cellcolor{green!20} 5.3 &  \cellcolor{green!20} 0.1 \\ 
        \hline
        \multirow{2}{1.4cm}{Max} & 2 beats 1 (\%)& 66.1 & 72.4 & 95.3 & 64.6 & 44.9 \\ 
        ~ & acc gain & \cellcolor{green!20}  3.0 & \cellcolor{green!20} 3.0 & \cellcolor{green!20} 4.4 & \cellcolor{green!20} 3.2 & \cellcolor{pink!50} -0.5 \\ 
        \hline
        \multirow{2}{1.4cm}{Average} & 2 beats 1 (\%)& 64.6 & 75.2 & 94.1 & 68.5 & 49.6 \\ 
        ~ & acc gain & \cellcolor{green!20} 2.8 & \cellcolor{green!20} 3.5 & \cellcolor{green!20} 4.3 & \cellcolor{green!20} 4.3 & \cellcolor{pink!50} -0.2 \\ 
        \bottomrule
    \end{tabular}
    }
    \caption{Comparison between single layer and layer aggregation settings. ``2 beats 1 (\%)'' refers to the percentage of cases where layer aggregation performance is at least as high as the single layer performance, under the same configuration (LM \& layer). ``Acc gain'' stands for the average accuracy gain of layer aggregation over single layer across all configurations. Positive accuracy gains are highlighted in green, negative ones in pink.}
    \label{table:single_layeragg_comparison}
\end{table}
\vspace{1mm}

\minisection{Configuration.} We experiment with two parameters in the configuration: LM and layer. Note that the purpose of experimenting with different configurations is not to solve the task, but rather to obtain a comprehensive picture of which embeddings best represent the target features.
\begin{itemize}
    \item \textbf{Language Models}: We experiment with both static and contextualized representations. For static embeddings, we consider GloVe \cite{pennington-etal-2014-glove} and fastText \cite{bojanowski-etal-2017-enriching}. For contextualized LMs, we consider encoder-only monolingual and multilingual Transformer models of different sizes (base and large), including BERT \cite{devlin-etal-2019-bert}, mBERT (multilingual BERT) \cite{devlin-etal-2019-bert}, RoBERTa \cite{liu2019roberta}, and XLM-RoBERTa \cite{conneau-etal-2020-unsupervised}.\footnote{See Appendix~\ref{sec:appendix_implementation} for implementation details.} 
    \item \textbf{Layer} ($l$): For contextualized LMs, another configuration choice is which layer to obtain the representation from. We explore the knowledge encoded in different layers in the range of 0-12 for base models and 0-24 for large ones, including the embedding layer.
\end{itemize}

\vspace{1mm}
\minisection{Pooling strategies.} In order to obtain a score for a feature of interest (complexity, formality, or figurativeness) for text segments that contain more than one token (i.e., phrases and sentences), 
we consider two pooling strategies over the scores calculated for individual tokens:\footnote{See Appendix~\ref{sec:appendix_implementation} for details on tokenization and multi-word expression handling.}

\begin{itemize}
  \item{\bf mean}: We compute the cosine similarity between $\overrightarrow{d_{feature}}$ and each word vector, and take the average of the similarity scores as the feature value for the text.
  \item {\bf max}: We compute the cosine similarity between $\overrightarrow{d_{feature}}$ and each word vector, and take the maximum of the similarity scores as the feature value for the text. 
\end{itemize}

The intuition behind max pooling is that the majority of words in a phrase or sentence would not be too extreme (i.e., too complex or too formal). By looking at the most complex or formal word in the text, we can get an idea of how extreme it might be in that dimension. Naturally, we expect this approach to perform less well than mean pooling for figurativeness, where idiomaticity is most often inferred by looking at the context of use and the word combinations within a sentence. 


\section{Results and Discussion}

Table~\ref{table:main_results} presents the results of our evaluation. 
Due to space constraints, each row in the table only shows the optimal performance obtained across all configurations (LM and layers) for static and contextualized models.\footnote{See Appendix~\ref{sec:appendix_lm_perf} for detailed accuracy scores for each model.}
For contextualized LMs in particular, following \citet{vulic-etal-2020-probing}, we separately show the optimal performance under two settings: \textbf{single layer}, where only the representation from a single layer $l$ is used; and \textbf{layer aggregation} (``layeragg'' for short), where we average the representations from all layers from the 0th to a specific layer $l$ (included). 

\begin{figure*}
  \centering
  
  \begin{subfigure}[b]{0.3\textwidth}
    \includegraphics[width=\textwidth]{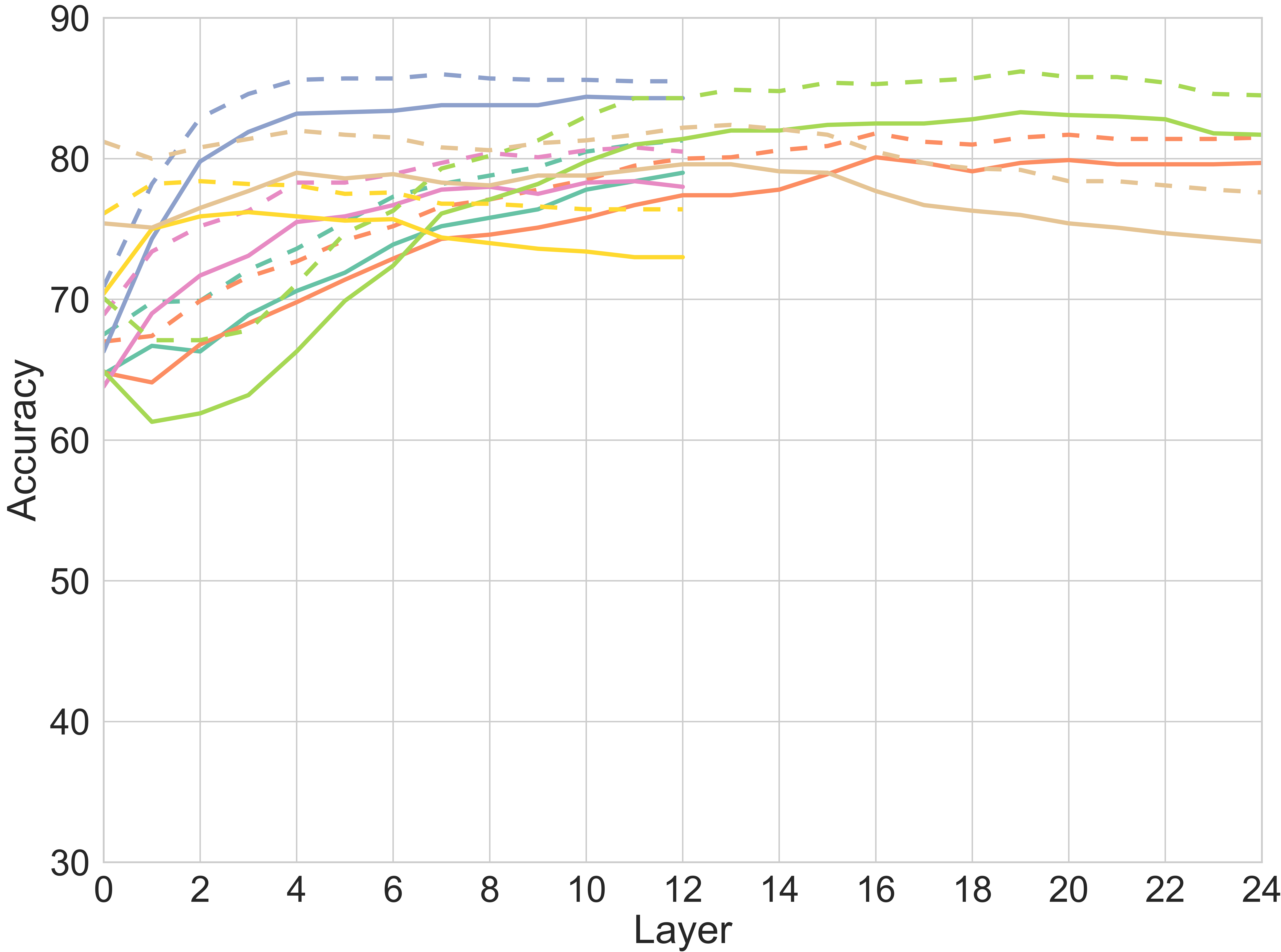}
    \caption{Complexity (short)}
    \label{fig:layer_complexity_short}
  \end{subfigure}
  \hfill 
  \begin{subfigure}[b]{0.3\textwidth}
    \includegraphics[width=\textwidth]{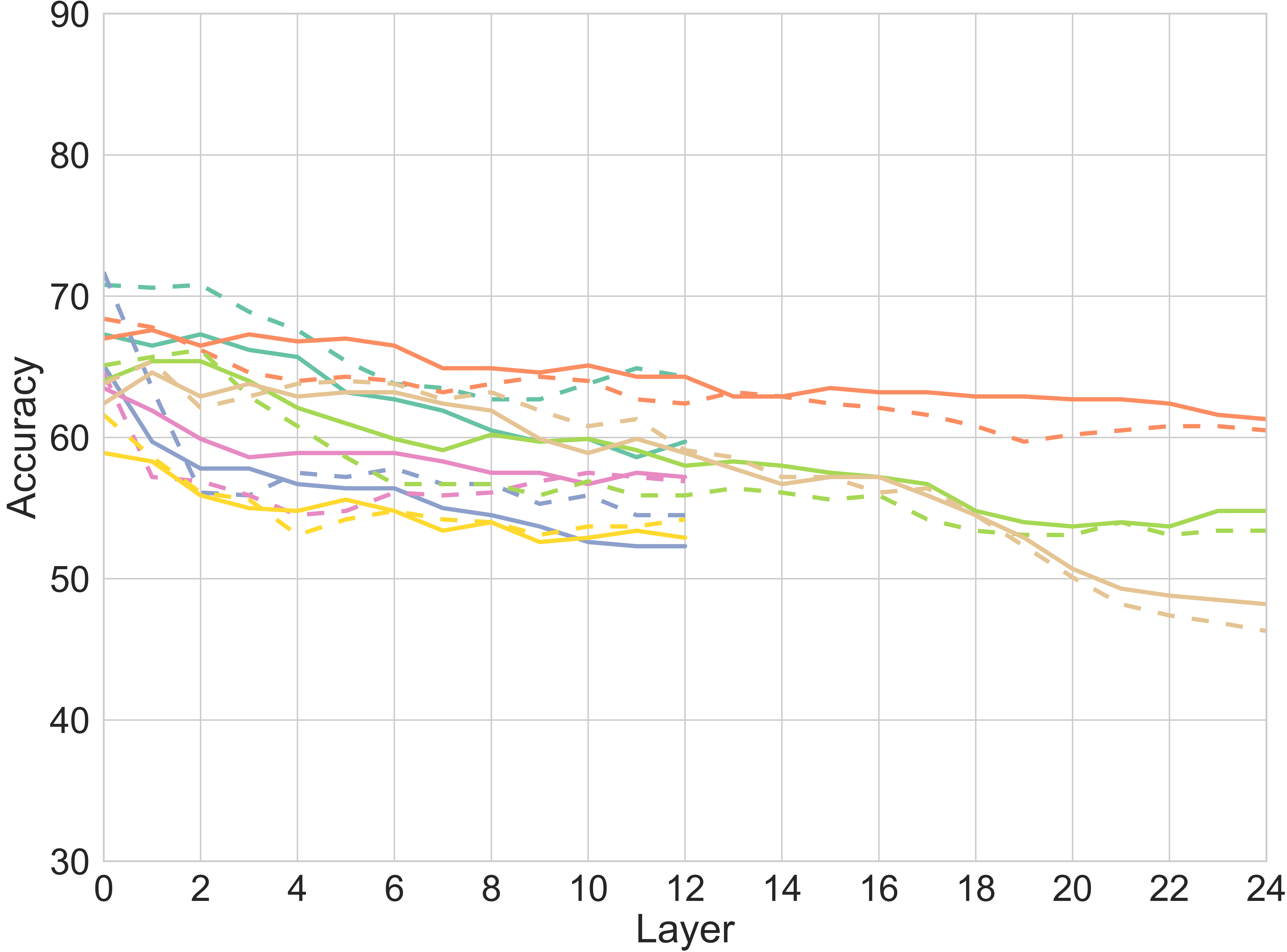}
    \caption{Formality (short)}
    \label{fig:layer_formality_short}
  \end{subfigure}
  \hfill 
  \begin{subfigure}[b]{0.3\textwidth}
    \raisebox{0.2cm}{\includegraphics[height=3.8cm]{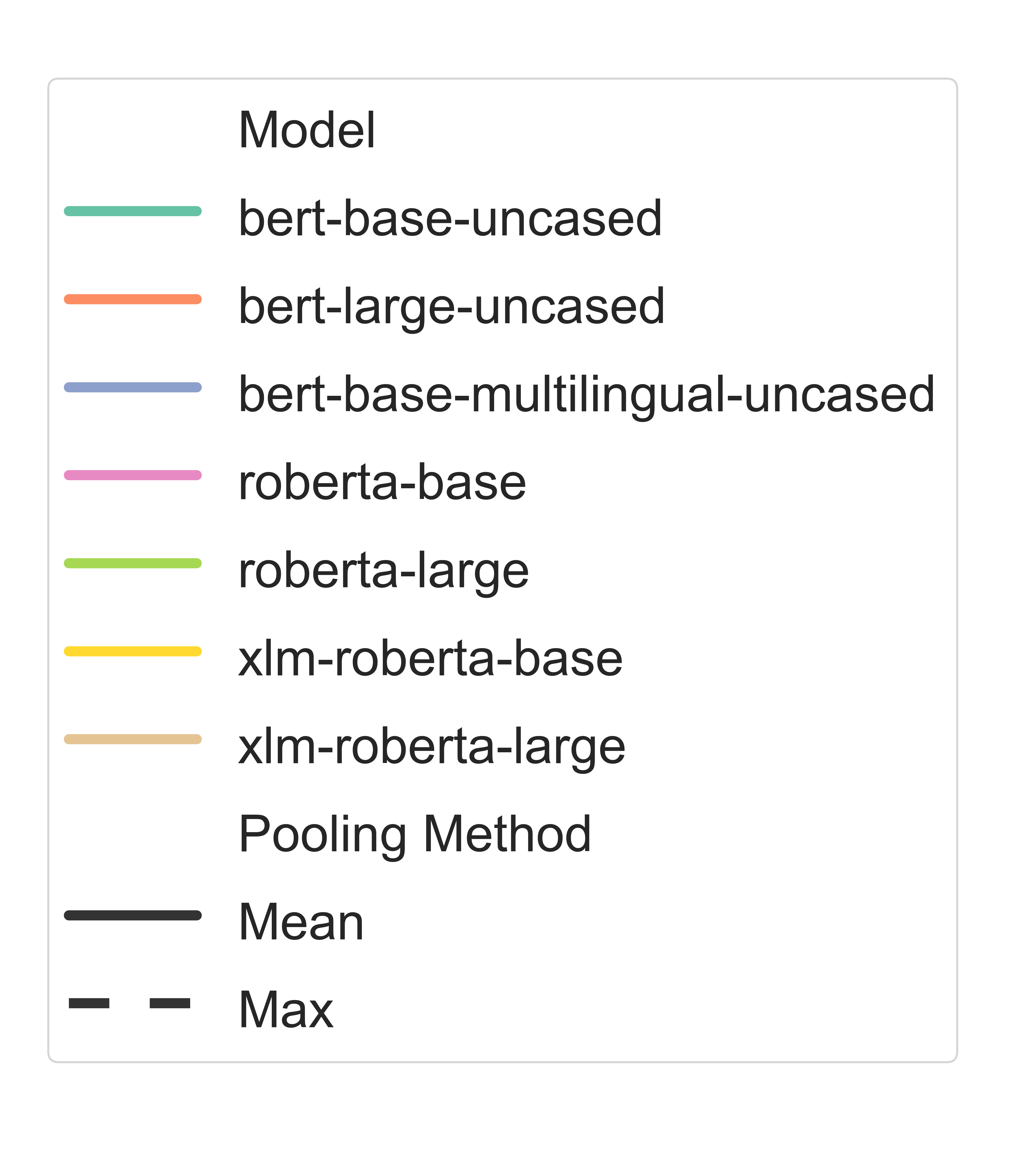}}
    \label{fig:sub4}
  \end{subfigure}
  
  \begin{subfigure}[b]{0.3\textwidth}
    \includegraphics[width=\textwidth]{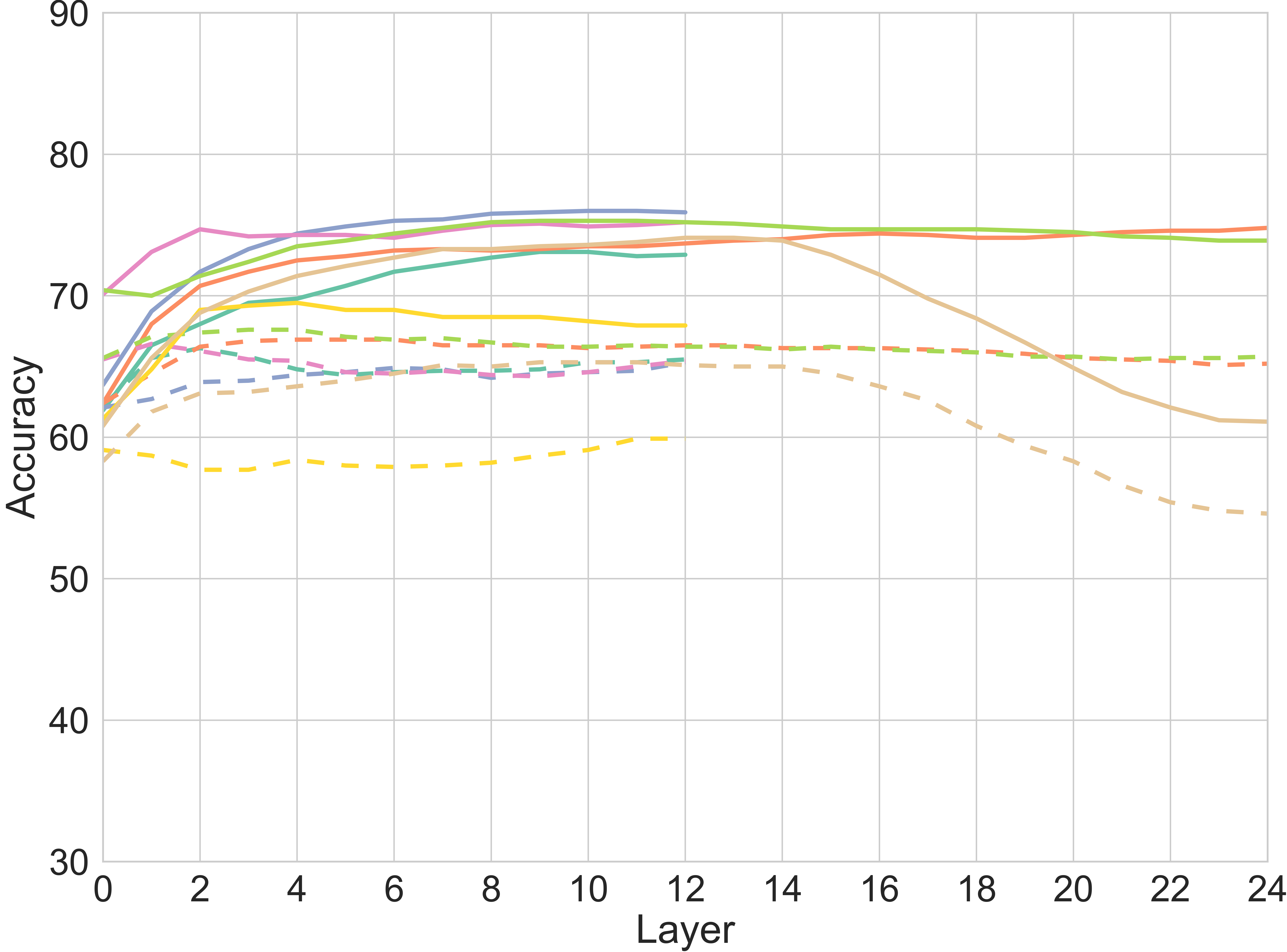}
    \caption{Complexity (long)}
    \label{fig:layer_complexity_long}
  \end{subfigure}
  \hfill 
  \begin{subfigure}[b]{0.3\textwidth}
    \includegraphics[width=\textwidth]{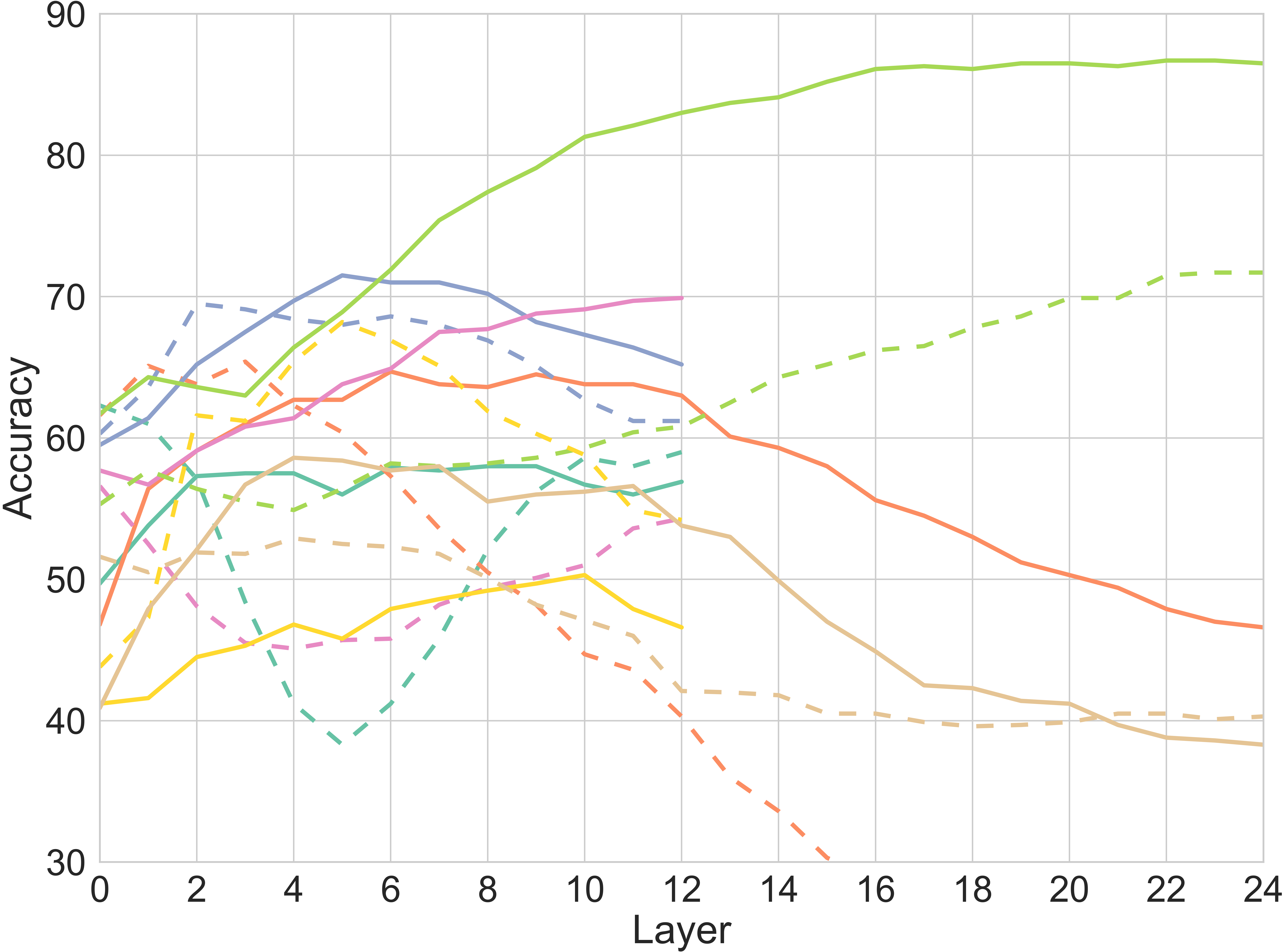}
    \caption{Formality (long)}
    \label{fig:layer_formality_long}
  \end{subfigure}
  \hfill
  \begin{subfigure}[b]{0.3\textwidth}
    \includegraphics[width=\textwidth]{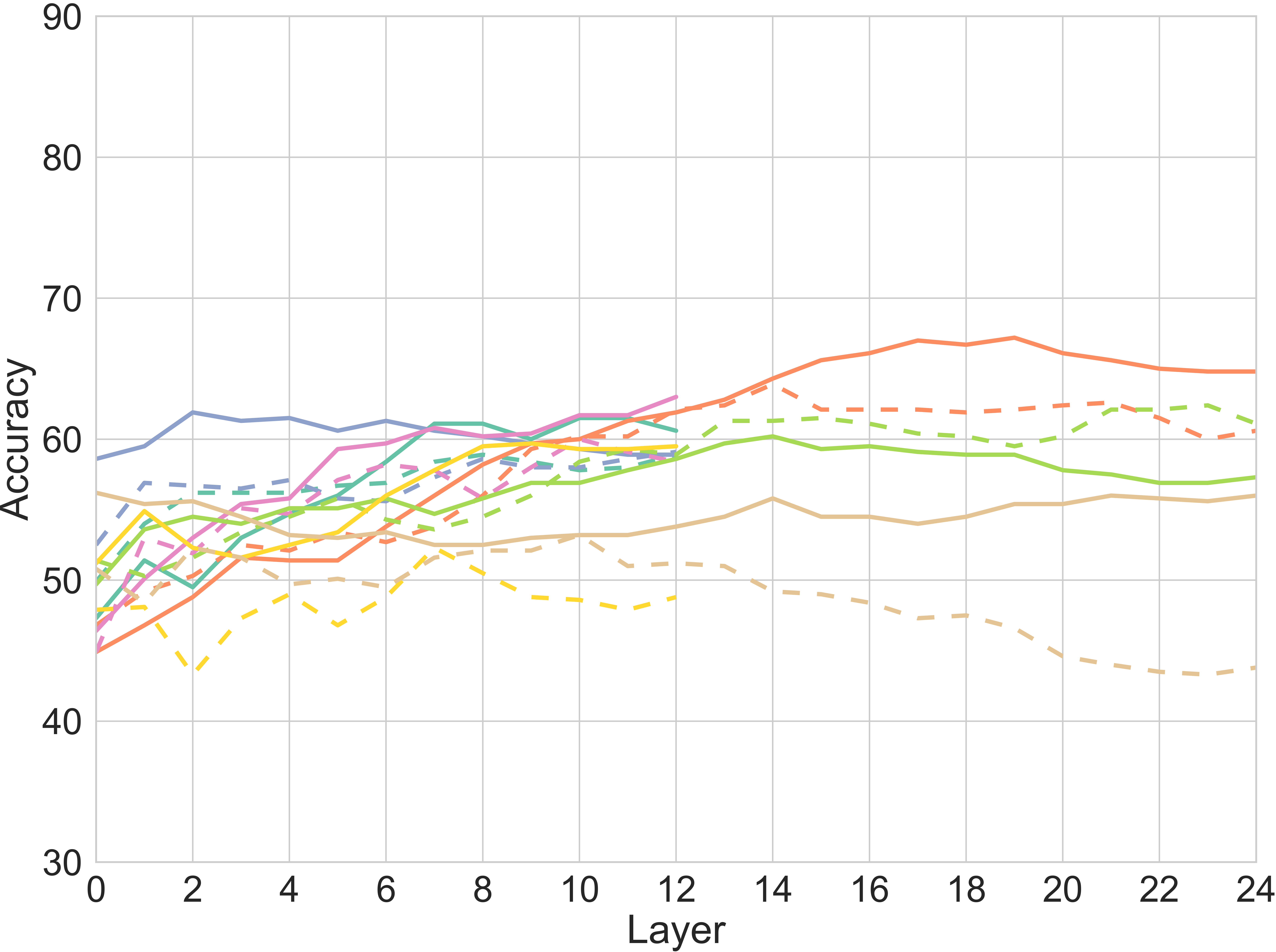}
    \caption{Figurativeness (long)}
    \label{fig:layer_figurativeness_long}
  \end{subfigure}
  \hfill 
  \caption{Performance change across layers of different LMs (under the layer aggregation setting). }
  \label{fig:layer_perf}
\end{figure*}

We observe that our method outperforms the majority and frequency baselines with both static and contextualized LMs. 
Furthermore, mean pooling generally works better than max pooling, although there is still room for improvement. Taking a closer look at the optimal configuration for each feature, for \textbf{complexity}, {\tt roberta-large} and {\tt mbert-base} are the dominant best-performing models, yet there are no consistently dominant layers; for \textbf{formality}, {\tt bert-base} and {\tt mbert-base} perform the best on short texts and surprisingly with the initial layers (0 or 1), while {\tt roberta-large} is the best model for long texts with middle or final layers; for \textbf{figurativeness}, {\tt bert-large} is consistently the best model across all settings.

Interestingly, contextualized LMs far outperform static embeddings on long text sin almost all cases, yet on short texts, static embeddings perform on par or sometimes even better than contextualized LMs. This is the case, for example, with formality ``short'' (with both pooling strategies) and with complexity ``short'' (with max pooling). This finding sounds counter-intuitive, given the generally higher performance of contextualized models in various NLP tasks. In our probing setting, we suspect that this might be due to two factors. First, the input in short-text datasets consists of isolated, rather than contextualized, instances of words. This is not natural input for a contextualized LM. Second, previous work has demonstrated that the word-level similarity estimates obtained from the vector space of contextualized LMs might be distorted due to the anisotropy of the space \cite{ethayarajh-2019-contextual,rajaee-pilehvar-2021-cluster}. Concretely, anisotropic word representations occupy a narrow cone instead of being uniformly distributed in the vector space, resulting in excessively positive correlations even for unrelated word instances. 
This has a negative impact on the informativeness of measures such as the cosine and the Euclidean distance, often used for estimating representation similarity \cite{10.1162/coli_a_00474}. 
These measures are dominated by a small subset of ``rogue dimensions'' which drive anisotropy and the drop in representational quality in later layers of the models \citep{timkey-van-schijndel-2021-bark}. 
In Section \ref{anisotropyreduction}, we investigate more closely the impact of anisotropy on our results through a series of experiments involving different anisotropy reduction methods. 

\begin{figure*}
  \centering
  
  \begin{subfigure}[b]{0.3\textwidth}
    \includegraphics[width=\textwidth]{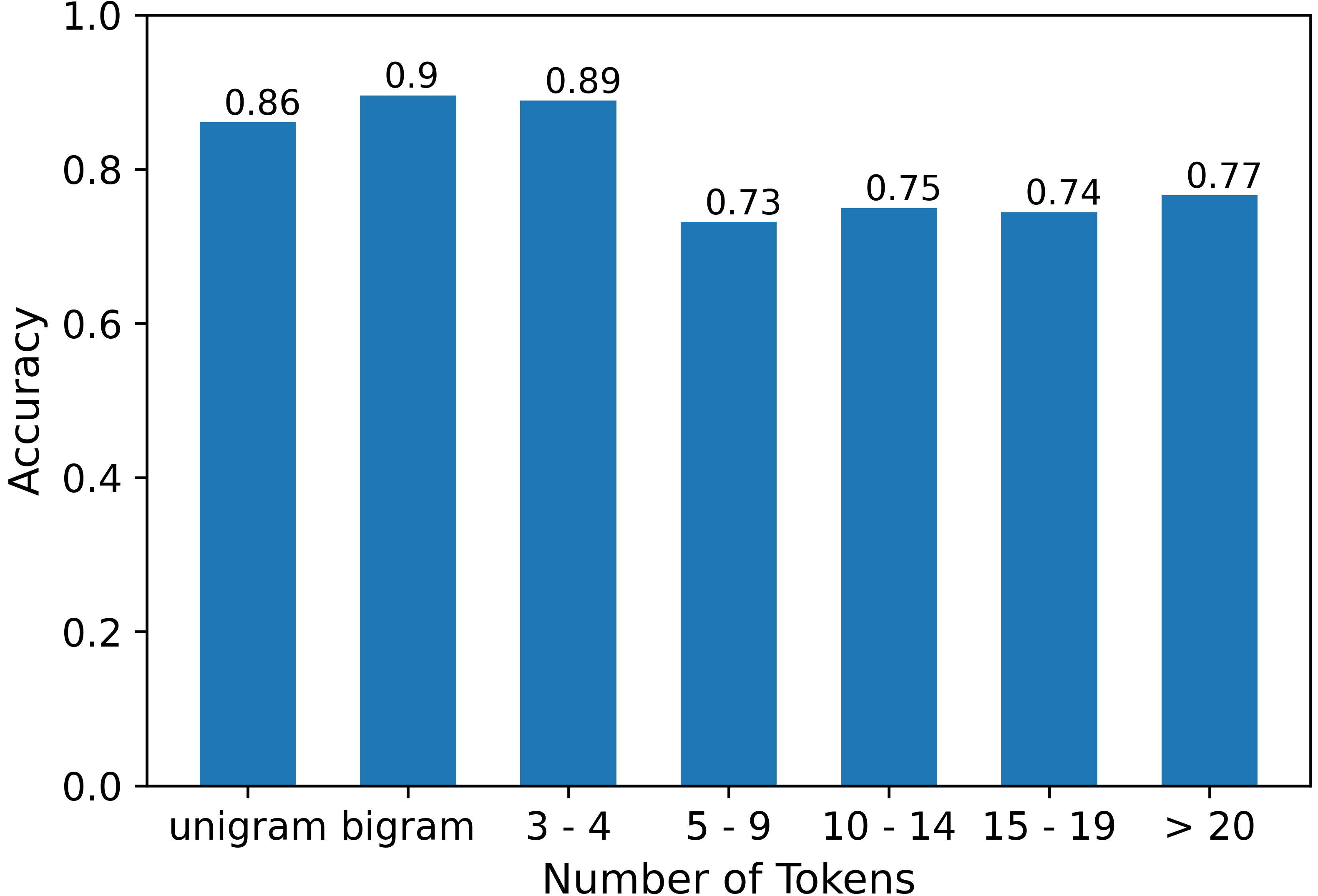}
    \caption{Complexity}
    \label{fig:length_complexity}
  \end{subfigure}
  \hfill 
  \begin{subfigure}[b]{0.3\textwidth}
    \includegraphics[width=\textwidth]{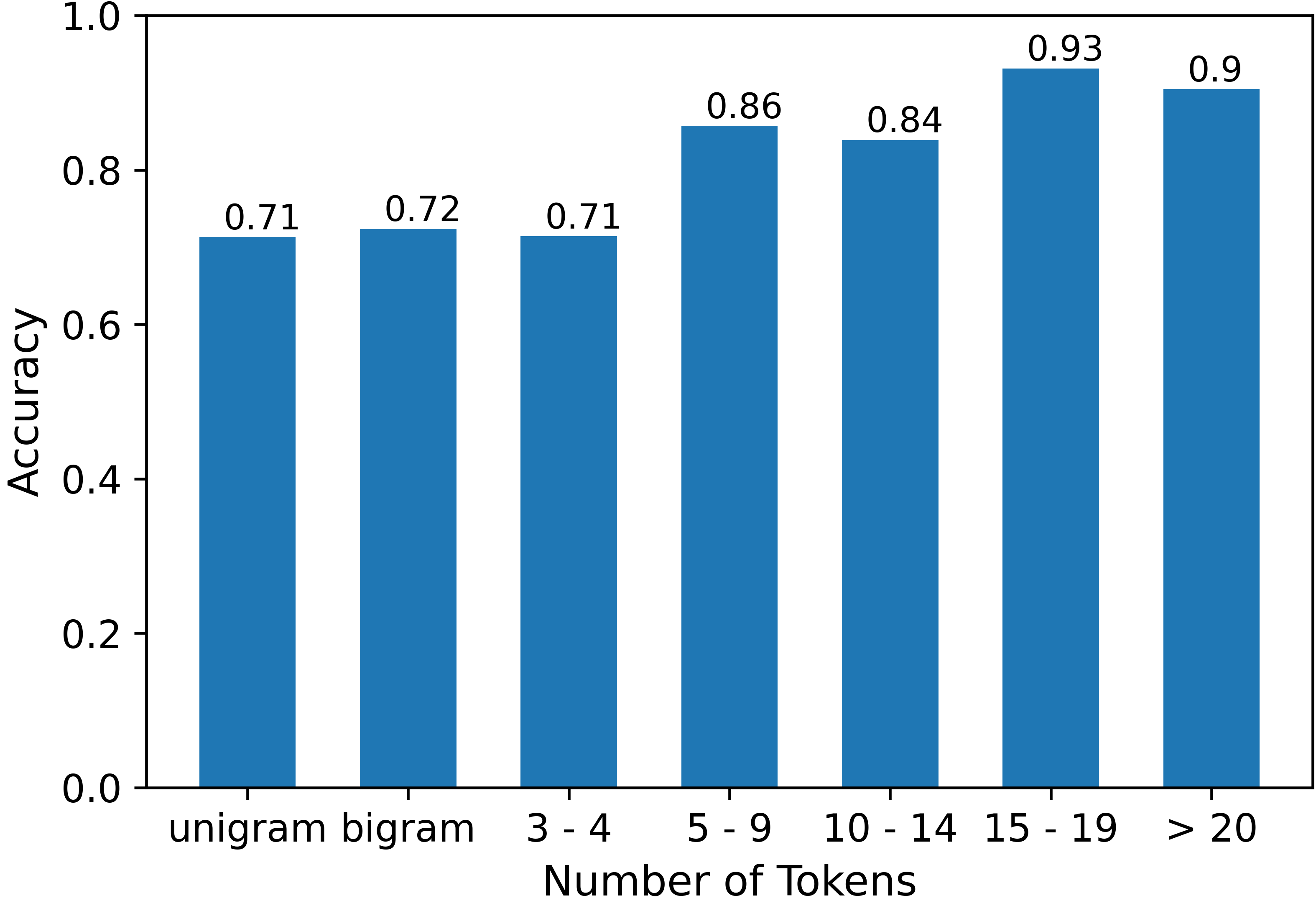}
    \caption{Formality}
    \label{fig:length_formality}
  \end{subfigure}
  \hfill 
  \begin{subfigure}[b]{0.3\textwidth}
    \includegraphics[width=\textwidth]{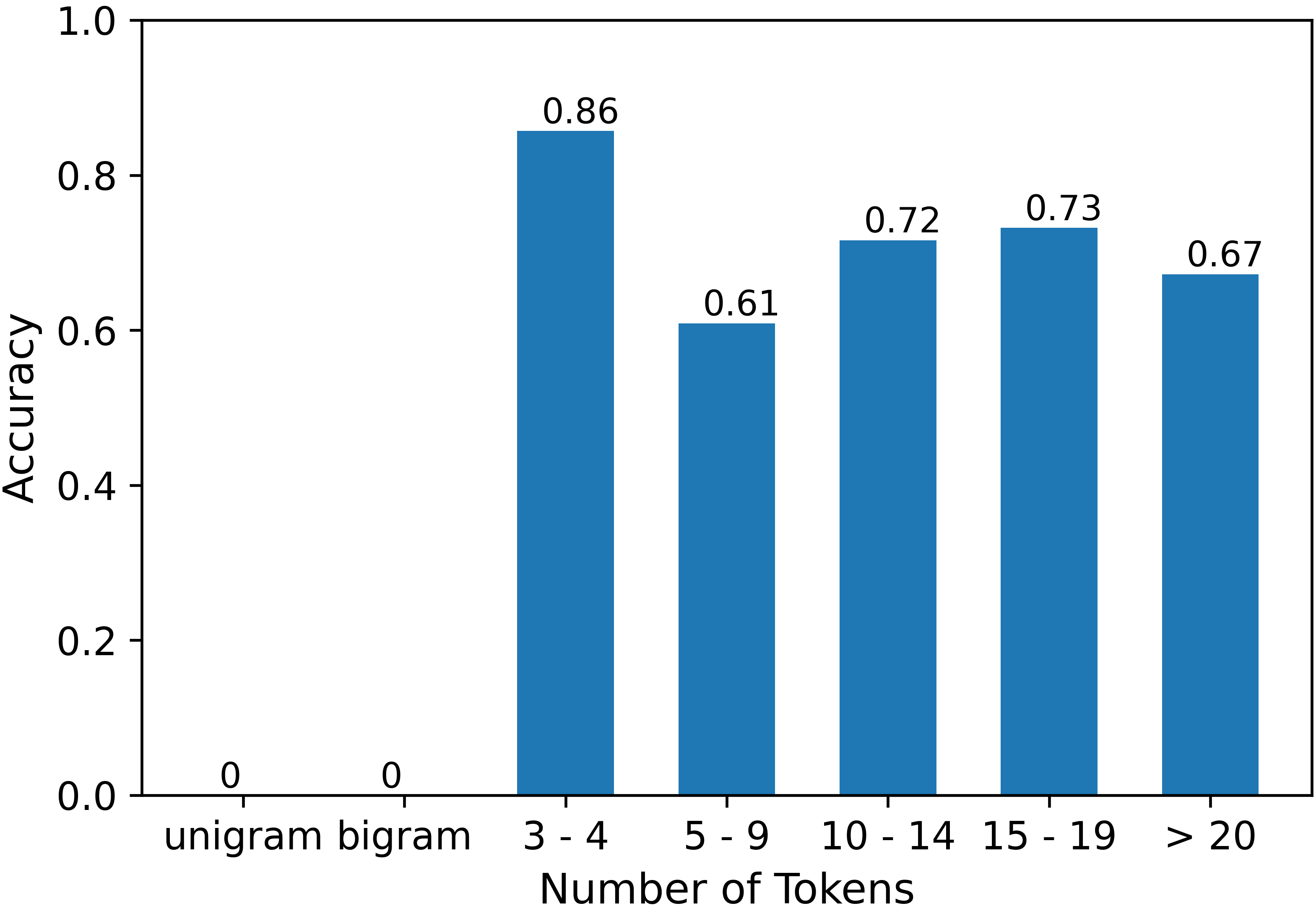}
    \caption{Figurativeness}
    \label{fig:length_figurativeness}
  \end{subfigure}
  
  \caption{Optimal performance over different bins of text length (under the layer aggregation setting).}
  \label{fig:acc_by_length_layeragg}
\end{figure*}

Finally, comparing the single layer and layer aggregation settings, their respective \textbf{optimal} configurations result in mostly similar performance across datasets, as shown in Table~\ref{table:main_results}. In order to better understand their difference across all possible LM and layer configurations, we present in Table \ref{table:single_layeragg_comparison} two types of averaged statistics: the percentage of configurations where the layer aggregation performance is equal or higher than the single layer performance, as well as the average gain in terms of accuracy. We observe that layer aggregation improves the performance for complexity and formality (across 64.6\% to 94.1\% of the configurations and by an accuracy gain of 2.8 to 4.3), but makes almost no difference for figurativeness. Together with the results from Table~\ref{table:main_results}, this suggests that although layer aggregation does not help with the best configuration, it is beneficial to most configurations on average.

In the next two subsections, we analyze the influence of two more factors on our method: layer depth and text length. For conciseness, we only report the results for the layer aggregation setting. Results for the single layer setting are given in Appendix~\ref{sec:appendix_extended_results}. 

\subsection{How well do different layers represent the target features?} 
\label{sec:perf_by_layer}

We explore the representation of the three stylistic features inside contextualized LMs by specifically monitoring the change in accuracy observed across layers. The results are shown in Figure \ref{fig:layer_perf}.
The solid curves show results obtained using mean pooling, while the dashed ones correspond to max pooling. 

We observe that information about \textbf{complexity} (\ref{fig:layer_complexity_short} and \ref{fig:layer_complexity_long}) is more clearly and consistently encoded after layer 4 of the models, independent of their size (base or large). Across all layers, mean and max pooling exhibit mostly similar behavior for short texts, while mean pooling is clearly better for longer texts. 
The pattern for \textbf{figurativeness} is similar (\ref{fig:layer_figurativeness_long}), though with slightly more fluctuations. For \textbf{formality}, we see a different trend. As shown in Figure~\ref{fig:layer_formality_short}, this feature is encoded more clearly in the early layers of the models for short texts. For longer texts, we see diverging patterns across layers of different models (\ref{fig:layer_formality_long}). In particular, {\tt roberta-large} encodes formality better than other tested models while its multilingual versions ({\tt xlm-roberta-base} and {\tt xlm-roberta-large}) give much lower results. 

\subsection{How does text length influence our method?} 
\label{sec:perf_by_length}

We analyze the performance change with regard to text length, represented by the average number of tokens in the two texts in a pair. For each feature, we merge examples from the short-text and long-text datasets (if available) and take the predictions from the best-performing contextualized configuration from Table~\ref{table:main_results}. Based on the number of tokens, we group all examples into several bins (unigram, bigram, 3-4, 5-9, 10-14, 15-19, >20) and compute the average accuracy in each bin. Figure~\ref{fig:acc_by_length_layeragg} shows the results. 

Interestingly, we observe different patterns for the three features. For \textbf{complexity}, accuracy scores for shorter texts (0.86 to 0.9) are generally higher than those for long texts (0.73 to 0.77). The drop from 3-4 tokens to 5-9 tokens is particularly clear. 
For \textbf{formality}, on the contrary, longer texts (0.84 to 0.93) tend to be easier than short ones (0.71 to 0.72). For \textbf{figurativeness}, we do not have results for short texts since no such datasets are available. Within full sentences, we observe that our method works better for shorter sentences (with <5 tokens) than for longer ones (with >=5 tokens) by an accuracy difference of 0.13 to 0.25. One caveat is that these differences are not only influenced by text length, but also by the intrinsic data distribution in different datasets. For example, the domain of the source texts in SimplePPDB (news, legal documents, and movie subtitles) is different from that in SimpleWikipedia (encyclopedia articles). Thus, the accuracy differences could be a result of both factors --- text length and text domain.

\section{Anisotropy Reduction Experiments}
\label{anisotropyreduction}

\begin{table*}[!t]
    \centering
    \scalebox{0.75}{
    \begin{tabular}{ll|cc|cc|c}
    \toprule
        \textbf{Pooling} & \textbf{Model} & \multicolumn{2}{c|}{\textbf{Complexity}} & \multicolumn{2}{c|}{\textbf{Formality}} & \textbf{Figurativeness} \\
        ~ & ~ & short & long & short & long & long \\
         \hline
        \multirow{9}{1.5cm}{Mean} & static & 84.8 & 60.0 & \textbf{76.8} & 82.8 & 54.3 \\ 
        ~ & contextualized (singlelayer) & 86.2 & \textbf{76.5} & 68.7 & 82.4 & \textbf{72.9} \\ 
        ~ & contextualized (singlelayer+\textbf{abtt}) & 80.3 & 69.3 & 76.6 & 76.7 & 70.9 \\ 
        ~ & contextualized (singlelayer+\textbf{standardization}) & \textbf{90.4} & 73.9 & 74.1 & 80.6 & 68.3 \\ 
        ~ & contextualized (singlelayer+\textbf{rank}) & 85.6 & 76.0 & 70.8 & 81.7 & 71.8 \\ 
        ~ & contextualized (layeragg) & 84.4 & 76.0 & 67.6 & \textbf{86.7} & 67.2 \\ 
        ~ & contextualized (layeragg+{\bf abtt}) & 81.7 & 68.5 & 76.6 & 63.0 & 72.6 \\ 
        ~ & contextualized (layeragg+{\bf standardization}) & \textbf{90.4} & 73.6 & 75.2 & 79.9 & 67.6 \\ 
        ~ & contextualized (layeragg+{\bf rank}) & 83.7 & 75.7 & 68.1 & 82.1 & 67.0 \\ 
        \hline
        \multirow{9}{1.5cm}{Max} & static & 89.4 & 58.0 & 76.0 & 63.4 & 56.0 \\ 
        ~ & contextualized (singlelayer) & 87.7 & 69.4 & 71.7 & 73.6 & 64.8 \\ 
        ~ & contextualized (singlelayer+\textbf{abtt}) & 80.6 & 64.9 & 78.2 & 80.8 & 66.7 \\ 
        ~ & contextualized (singlelayer+\textbf{standardization}) & \textbf{90.5} & 63.8 & \textbf{80.9} & \textbf{81.7} & 60.4 \\ 
        ~ & contextualized (singlelayer+\textbf{rank}) & 87.1 & \textbf{69.6} & 70.3 & 76.0 & 66.5\\
        ~ & contextualized (layeragg) & 86.2 & 67.6 & 71.7 & 71.7 & 63.9 \\ 
        ~ & contextualized (layeragg+{\bf abtt}) & 81.9 & 63.9 & 78.2 & 72.5 & \textbf{71.1} \\ 
        ~ & contextualized (layeragg+{\bf standardization}) & \textbf{90.5} & 63.7 & \textbf{80.9} & 80.6 & 61.9 \\ 
        ~ & contextualized (layeragg+{\bf rank}) & 86.1 & 69.3 & 71.7 & 71.5 & 67.4 \\ 
        \bottomrule
    \end{tabular}
    }
    \caption{Performance of three anisotropy reduction methods (all-but-the-top/standardization/rank-based). The highest performance within each pooling method (Mean/Max) is in boldface.}
    \label{table:isotropy}
\end{table*}

As explained in the previous section, the anisotropy of contextualized LMs' representation space degrades the quality of the similarity estimates that can be drawn from it \cite{ethayarajh-2019-contextual}.
To see if this has an impact on our method, we apply three post-processing anisotropy reduction methods discussed by \citet{timkey-van-schijndel-2021-bark}, which can be used to correct for rogue dimensions and reveal underlying representational quality. 

We apply each of these methods to our feature vector construction and feature value prediction processes. 
Given that our stylistic characterization of new text relies on similarity measurement, we expect that a space that allows us to draw higher-quality similarity estimates would better represent these stylistic features and would also improve feature value prediction. 
The three methods used in our experiments are:

\vspace{1mm}
\minisection{All-but-the-top (abtt)}. The method was initially proposed for static embeddings by \citet{mu2018allbutthetop}. The main idea is to subtract the common mean vector and eliminate the top few principal components (PCs) (we use the top $\frac{d}{100}$, where $d$ represents the dimensionality of the vector space, following their suggestion). These subtracted vectors should capture the variance of the rogue dimensions in the model and make the space more isotropic. In \citet{timkey-van-schijndel-2021-bark}, the mean vector and PCs are computed from vector representations for an entire corpus. Since our method is unsupervised, we do not assume access to any large corpus and instead compute them based only on the seed pairs (i.e., 14 words and phrases for each feature). Thus, our method still remains lightweight and computationally efficient. It is, however, important to note that this is a {\bf local correction} (rather than a global one) since we are just using a small number of words and phrases, as in \citet{rajaee-pilehvar-2021-cluster}. 

Formally, given a set of seed texts of size $|\mathcal{S}|$ (here $|\mathcal{S}|=14$) containing token representations $x \in \mathbb{R}^d$, we compute the mean vector $\mu \in \mathbb{R}^d$
\begin{equation}
\label{eq:mean_vec}
\mu = \frac{1}{|\mathcal{S}|} \cdot \sum_{x \in \mathcal{S}} x
\end{equation}

\noindent as well as the PCs
\begin{equation}
u_1,...,u_d = {\rm PCA}(\{x-\mu, x \in \mathcal{S}\}).
\end{equation}

\noindent Then, the new representation $x_{abtt}$ for an unseen word vector $x$ is the result of eliminating the mean vector and the top $k$ PCs (here $k=\frac{d}{100}$):

\begin{equation}
x_{abtt} = x - \mu - \sum_{i=1}^k \left(u_i^{\top} x\right) u_i. \\
\end{equation}

\vspace{1mm}
\minisection{Standardization}. Based on 
a similar observation as {\bf abtt} (a non-zero common mean vector and a few dominant directions), another way for adjustment is to subtract the mean vector and divide each dimension by its standard deviation (std), such that each dimension has $\mu_{i} = 0$ and $\sigma_i = 1$. Similarly to {\bf abtt}, we compute the mean vector and standard deviation using only the seed pairs for each feature.

Formally, we compute the same mean vector $\mu$ as in Equation~\ref{eq:mean_vec}, as well as the standard deviation in each dimension $\sigma \in \mathbb{R}^d$
\begin{equation}
\sigma = \sqrt{\frac{1}{|\mathcal{S}|} \cdot \sum_{x \in \mathcal{S}} (x - \mu)^2}
\end{equation}

\noindent The new representation $x_{standard}$ for an unseen word vector $x$ becomes
\begin{equation}
x_{standard} = \frac{x - \mu}{\sigma}.
\end{equation}

\minisection{Rank-based}. 
This method treats a word vector as $d$ observations from an $|\mathcal{S}|$-variate distribution and uses correlation metrics as a measure of similarity, instead of cosine similarity \cite{zhelezniak-etal-2019-correlation}. Specifically, Spearman's $\rho$, a non-parametric correlation measure, only considers the ranks of embeddings rather than their values. Thus, it will not be dominated by the rogue dimensions of contextualized LMs. Unlike the previous two methods, this method does not require any computation over the seed pair texts. Formally, given a word vector $x$, the new representation $x_{rank}$ is simply

\begin{equation}
 x_{rank} = rank(x).
 \vspace{-0.2in}
\end{equation}\\

Table~\ref{table:isotropy} shows the effect of applying the three anisotropy reduction strategies under the single layer and layer aggregation settings. Overall, after anisotropy reduction, contextualized LMs outperform static embeddings in all cases except formality ``short'', confirming our initial hypothesis. Nevertheless, there is no universally optimal strategy, although standardization works best most of the time. Comparing the two pooling strategies, we find that anisotropy correction helps more often with max pooling than with mean pooling. 

It is important to reemphasize that our anisotropy correction approach is local, since it only considers a small set of words and phrases for calculating the mean vector, standard deviation, and PCs. This might be the reason for the relatively small observed effect of these correction procedures in our experiments. In future work, we plan to experiment with a larger corpus, and consequently use a larger part of the vector space for calculating the mean/std/PC vectors, in order to investigate the impact of the quantity of data on the induced similarity estimates.

\section{Conclusion}

We have shown that the embedding space of pretrained LMs encodes abstract stylistic notions such as formality, complexity, and figurativeness. Using a geometry-based method, we construct a vector representation for each of these features, which can be used to characterize new texts. We find that these notions are present in the space of both static and contextualized representations, and that static embeddings are better at capturing the style of short texts (words and phrases) whereas contextual embeddings at longer texts (sentences). By correcting the anisotropy of contextualized LMs' representation space, we show that it is possible to close the performance gap from static embeddings on short texts. 

Our unsupervised and lightweight method is expected to be applicable for stylistic analysis in other languages and for other stylistic notions, such as concreteness, sentiment, and political stance, which we plan to address in future work. Furthermore, we plan to experiment with anisotropy correction methods on a larger corpus, and to adapt the method for style prediction on longer text (e.g., whole documents). The stylistic measurements obtained using this method can be useful in the creation of lexical style lexicons as well as in downstream applications, for authorship attribution and style transfer.

\section*{Limitations}

We acknowledge the following limitations of our work: (a) The scope of our experiments is limited to the English language currently. Our method is only evaluated on the level of words, phrases, and sentences, but not at the document level. (b) The effect of anisotropy reduction strategies is shown to be rather mixed. Further investigation is required to determine under what conditions these strategies can prove beneficial in the specific context of stylistic feature extraction. (c) Our work addresses only lexical-level stylistic features and not more global aspects of writing style, such as the diversity of word choice and the utilization of unique syntactic structures. Whether this method can be extended to capture the comprehensive nuances of writing style is an interesting direction for future work.

\section*{Ethical Considerations}

In this paper, our method is only tested in intrinsic evaluation settings where existing publicly available datasets have been used. It is not integrated into any downstream application, although this type of stylistic analysis could be potentially useful in different settings. 



\section*{Acknowledgments}
This research is based upon work supported in part by the DARPA KAIROS Program (contract FA8750-19-2-1004), the DARPA LwLL Program (contract FA8750-19-2-0201), the IARPA HIATUS Program (contract 2022-22072200005), and the NSF (Award 1928631). Approved for Public Release, Distribution Unlimited. The views and conclusions contained herein are those of the authors and should not be interpreted as necessarily representing the official policies, either expressed or implied, of DARPA, IARPA, NSF, or the U.S. Government.

\bibliography{custom}
\bibliographystyle{acl_natbib}

\newpage
\appendix

\section{Dataset Details}
\label{sec:appendix_dataset}

All evaluation datasets we use contain 
semantically similar but stylistically different words, phrases, or sentences. 

\subsection{Data Description and Source}

\paragraph{Complexity}
\begin{itemize}
  \item SimplePPDB \cite{pavlick-callison-burch-2016-simple}: It contains 4.5M pairs of words and short phrases, where one is simpler and the other is more complex. It is constructed based on a subset of the Paraphrase Database (PPDB) \cite{ganitkevitch-etal-2013-ppdb}. There are both automatically generated and manually annotated pairs.\footnote{For all datasets, we only use a subset of all pairs based on quality filtering, which is described in Appendix~\ref{sec:appendix_preprocessing}.} URL: \url{http://www.seas.upenn.edu/~nlp/resources/simple-ppdb.tgz}.
  \item SimpleWikipedia \cite{kauchak-2013-improving}: It contains 167K pairs of simple/complex sentences generated by aligning Simple English Wikipedia and English Wikipedia. We are using Version 2.0 of the dataset (updated from Wikipedia pages downloaded in May 2011), the ``Sentence-aligned'' subset. URL: \url{https://cs.pomona.edu/~dkauchak/simplification/data.v2/sentence-aligned.v2.tar.gz}.
\end{itemize}

\paragraph{Formality}
\begin{itemize}
  \item StylePPDB \cite{pavlick-nenkova-2015-inducing}: It contains 4.9K pairs of casual/formal words or short phrases from PPDB, 
  both automatically generated and manually annotated. URL: \url{https://cs.brown.edu/people/epavlick/data.html#style-pp-bibtex}.
  \item GYAFC \cite{rao-tetreault-2018-dear}: It contains a total of 110K informal/formal sentence pairs, created using the Yahoo Answers corpus. \footnote{\url{https://webscope.sandbox.yahoo.com/catalog.php?datatype=l&did=11}}. URL: \url{https://github.com/raosudha89/GYAFC-corpus}.
\end{itemize}

\paragraph{Figurativeness}
\begin{itemize}
  \item IMPLI \cite{stowe-etal-2022-impli}: It consists of 25.8K literal/figurative sentence pairs, spanning idioms and metaphors, both semi-supervised and human-annotated. URL: \url{https://github.com/UKPLab/acl2022-impli}.
\end{itemize}

\subsection{Preprocessing Method}
\label{sec:appendix_preprocessing}

To reduce noise and construct splits, we preprocess the datasets as follows:

\begin{itemize}
  \item SimplePPDB: There are both automatically and manually labeled subsets. We only take the manually labeled examples with $\geq 80\%$ of annotators agreeing with the final label. There are only training and validation sets in the original dataset. Since our method requires no training, we take the original training set as our validation set, and the original validation set as our test set.

  \item SimpleWikipedia: Since our method focuses on complexity in terms of lexical choice but not grammatical structure, we filter out pairs where the two sentences share the exact same set of tokens, or all tokens in a sentence appear in the other sentence. As there are no official splits, we randomly split the filtered dataset into train/validation/test sets of ratio 8:1:1 (since the dataset is huge).

  \item StylePPDB: The filtering method is the same as that used for SimplePPDB. There are no official splits either, so we randomly split the filtered dataset into a validation set and a test set of the same size (since the dataset is small).

  \item GYAFC: We take the Entertainment \& Music subset, using pairs from the files \texttt{formal} and \texttt{informal.ref0}. Since the official splits only have training and test sets, we take only the test set and re-split it into a new validation set and test set of the same size.

  \item IMPLI: We take the \texttt{manual\_e} subsets (manually created, entailing) for both idioms and metaphors, combine them and re-split the examples into a validation set and test set of the same size.

\end{itemize}

Finally, we randomly re-assign the label of every example for class balance.

\subsection{Statistics and Examples}

\begin{figure}[t!]
    \centering
    \includegraphics[width=0.9\columnwidth]{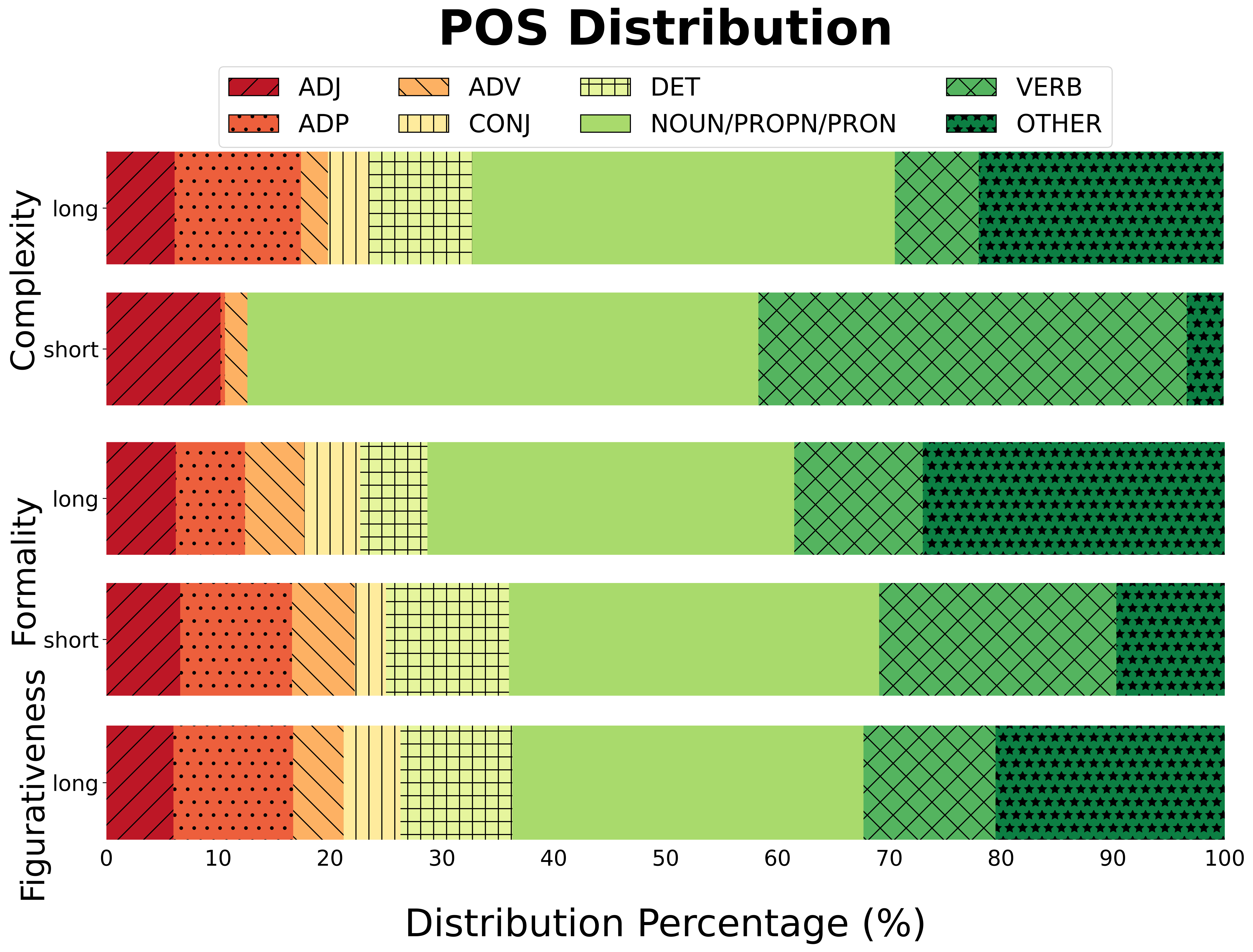}
    \caption{Distribution of token POS in evaluation datasets.}
    \label{fig:POS_stats}
\end{figure}

Table~\ref{table:dataset_details} shows the dataset statistics and example inputs and outputs after our preprocessing. Table~\ref{fig:POS_stats} shows the POS distribution statistics.

\begin{table*}[t]
\centering
\scalebox{0.8}{
\begin{tabular}{p{2cm}p{2.5cm}>{\raggedright\arraybackslash}p{1.5cm}>{\raggedright\arraybackslash}p{1.5cm}p{10cm}}
    \hline \textbf{Feature} & \textbf{Dataset} & \bf{\# Val}  & \bf{\# Test} & \textbf{Example}  \\
    \hline 
    \multirow{9}{2cm}{Complexity} & \multirow{3}{2cm}{SimplePPDB \newline (short)} & \multirow{3}{1.5cm}{814} & \multirow{3}{1.5cm}{1,108} & Text 0: \underline{toys} \newline Text 1: \underline{playthings} \newline Answer: \texttt{1} (more complex) \\ \cdashline{2-5} 
     & \multirow{6}{2cm}{SimpleWikipedia \newline (long) } & \multirow{6}{1.5cm}{9,978} & \multirow{6}{1.5cm}{9,978} & Text 0: Endemic types or species are especially likely to develop on biologically isolated areas such as islands \underline{because of their geographical isolation}.
 \newline Text 1: Endemic types are most likely to develop on islands \underline{because they are isolated}. \newline Answer: \texttt{0} (more complex)  \\
    \hline
    \multirow{6}{2cm}{Formality} & \multirow{3}{2cm}{StylePPDB \newline (short)} & \multirow{3}{1.5cm}{367} & \multirow{3}{1.5cm}{367} & Text 0: \underline{are allowed to} \newline Text 1: \underline{can} \newline Answer: \texttt{0} (more formal) \\ \cdashline{2-5}
     & \multirow{3}{1.8cm}{GYAFC \newline (long)} & \multirow{3}{1.5cm}{541} & \multirow{3}{1.5cm}{541}  & Text 0: \underline{I am impatiently waiting} to ask my husband\underline{.} \newline Text 1: \underline{Can't wait} to ask my husband\underline{!!} \newline Answer: \texttt{0} (more formal) \\    
    \hline
    \multirow{3}{2cm}{Figurativeness} & \multirow{3}{2cm}{IMPLI \newline (long)} & \multirow{3}{1.5cm}{243} & \multirow{3}{1.5cm}{243} & Text 0: You must \underline{adhere to} the rules.	\newline Text 1: You must \underline{obey} the rules. \newline Answer: \texttt{0} (more figurative) \\
    \hline
\end{tabular}
}
\caption{Datasets used for evaluation. ``\# Val'' and ``\# Test'' stand for the number of examples in the validation set and the test set respectively. Differences between pairs are underlined.}
\label{table:dataset_details}
 \vspace{-0.1in}
\end{table*}

\section{Implementation Details}
\label{sec:appendix_implementation}

\subsection{Tokenization}

Given a piece of text, we tokenize it with the SpaCy tokenzier\footnote{\url{https://spacy.io/api/tokenizer}} into words. Then, using the method described in \ref{sec:method}, we obtain a score for the feature of interest for each word token (if using static embeddings) or subword tokens (if using contextualized embeddings with WordPiece tokenization). In the latter case, we additionally obtain an aggregated feature score for each word from the scores of its subword tokens using a pooling strategy described in Section~\ref{sec:exp_setup}. Finally, we obtain an overall feature score for the entire piece of text from the scores of all its words using the same pooling strategy.

\subsection{Representations}

We use the following static embeddings: for GloVe, we use \texttt{GloVe.6B.300d}\footnote{\url{https://nlp.stanford.edu/projects/glove/}}, consisting of 400K word vectors trained on Wikipedia 2014 and Gigaword 5; and for fastText, we use \texttt{wiki-news-300d-1M-subword}\footnote{\url{https://fasttext.cc/docs/en/english-vectors.html}}, consisting of 1 million word vectors trained with subword infomation on Wikipedia 2017, UMBC webbase corpus and statmt.org news dataset. Out-of-Vocabulary (OOV) tokens are represented with the all-zero vector.

For contexutalized LMs, we use the following pretrained model checkpoints from HuggingFace Transformers\footnote{\url{https://github.com/huggingface/transformers}}: \texttt{bert-base-uncased} (110M parameters), \texttt{bert-large-uncased} (336M parameters), \texttt{bert-base-multilingual-uncased} (110M parameters), \texttt{roberta-base} (125M parameters), \texttt{roberta-large} (335M parameters), \texttt{xlm-roberta-base} (~125M parameters), \texttt{xlm-roberta-large} (~335M parameters). 

\subsection{Experiments}

We perform grid search on hyperparameters including the LM and the layer (0-12 for base models, and 0-24 for large models) using the validation set and report the performance of the optimal configuration on the test set. The optimal hyperparameters can be found in Appendix~\ref{sec:appendix_lm_perf}. 

All evaluation experiments are run on a single NVIDIA GeForce RTX 2080 Ti GPU node. Each experiment takes approximately 2-20 minutes depending on the size of the dataset.

\section{Extended Results}
\label{sec:appendix_extended_results}

In this section, we present additional results that cannot fit into Section~\ref{table:main_results} due to space limit.

\subsection{Performance of Different LMs}
\label{sec:appendix_lm_perf}

\begin{table*}[h]
    \centering
    \scalebox{0.8}{

    \begin{tabular}{ll|ll|ll|l}
    \toprule
        \textbf{Pooling} & \textbf{Model} & \multicolumn{2}{c|}{\textbf{Complexity}} & \multicolumn{2}{c|}{\textbf{Formality}} & \textbf{Figurativeness} \\
        ~ & ~ & short & long & short & long & long \\
         \hline
        ~ & majority & 55.1 & 50.6 & 51.2 & 51.8 & 51.4 \\
        \hline
        \multirow{10}{1.5cm}{Mean} & frequency & 83.2 & 51.0 & 61.0 & 41.4 & 49.7 \\ 
        ~ & fasttext.wiki & 73.1 & 58.4 & 61.6 & 45.1 & 52.7 \\ 
        ~ & glove.6B.300d & 84.8 & 60.0 & \textbf{76.8} & \textbf{82.8} & 54.3 \\         
        ~ & bert-base-uncased & 82.0 (10) & 72.2 (6) & 68.7 (1) & 63.2 (7) & 66.5 (10) \\ 
        ~ & bert-large-uncased & 82.3 (14) & 73.0 (5) & 68.1 (1) & 67.7 (3) & \textbf{72.9 (14)} \\ 
        ~ & bert-base-multilingual-uncased & 83.8 (1) & \textbf{76.5} (4) & 65.1 (0) & 72.1 (5) & 61.5 (4) \\ 
        ~ & roberta-base & 85.2 (4) & 75.5 (12) & 63.5 (0) & 76.7 (12) & 64.1 (3) \\ 
        ~ & roberta-large & \textbf{86.2} (4) & 75.3 (4) & 64.3 (1) & 82.4 (12) & 63.9 (4) \\ 
        ~ & xlm-roberta-base & 74.8 (4) & 69.6 (6) & 58.9 (0) & 56.4 (6) & 66.3 (6) \\ 
        ~ & xlm-roberta-large & 85.8 (11) & 73.7 (6) & 62.4 (0) & 67.7 (3) & 60.6 (23) \\ 
        
        \hline
        \multirow{10}{1.5cm}{Max} & frequency & 80.7 & 46.4 & 57.2 & 42.5 & 47.9 \\ 
        ~ & fasttext.wiki & 82.0 & 54.3 & 74.9 & 47.7 & 56.0 \\ 
        ~ & glove.6B.300d & \textbf{89.4} & 58.0 & \textbf{76.0} & 63.4 & 55.8 \\         
        ~ & bert-base-uncased & 83.7 (10) & 69.1 (12) & 70.8 (1) & 70.8 (8) & 64.6 (11) \\ 
        ~ & bert-large-uncased & 83.0 (6) & 67.6 (24) & 68.9 (1) & 64.1 (1) & \textbf{64.8} (11) \\ 
        ~ & bert-base-multilingual-uncased & 85.6 (1) & 65.7 (3) & 71.7 (0) & \textbf{73.6} (1) & 60.8 (8) \\ 
        ~ & roberta-base & 85.9 (4) & \textbf{69.4} (12) & 64.6 (0) & 70.1 (11) & 62.8 (5) \\ 
        ~ & roberta-large & 87.7 (4) & 68.9 (24) & 65.1 (0) & 72.1 (21) & 63.2 (6) \\ 
        ~ & xlm-roberta-base & 77.3 (1) & 64.3 (11) & 61.6 (0) & 70.2 (5) & 55.1 (11) \\ 
        ~ & xlm-roberta-large & 87.0 (11) & 67.1 (6) & 63.8 (0) & 61.4 (24) & 55.8 (3) \\         
        \bottomrule
    \end{tabular}
    }
    \caption{Accuracy of different models under the \textbf{single-layer} setting. The optimal layer number for each contextualized LM is in brackets. The highest performance within each pooling method is in boldface.}
    \label{table:lm_perf_singlelayer}
\end{table*}
\begin{table*}[h]
    \centering
    \scalebox{0.8}{

    \begin{tabular}{ll|ll|ll|l}
    \toprule
        \textbf{Pooling} & \textbf{Model} & \multicolumn{2}{c|}{\textbf{Complexity}} & \multicolumn{2}{c|}{\textbf{Formality}} & \textbf{Figurativeness} \\
        ~ & ~ & short & long & short & long & long \\
         \hline
        ~ & majority & 55.1 & 50.6 & 51.2 & 51.8 & 51.4 \\
        \hline
        \multirow{10}{1.5cm}{Mean} & frequency & 83.2 & 51.0 & 61.0 & 41.4 & 49.7 \\ 
        ~ & fasttext.wiki & 73.1 & 58.4 & 61.6 & 45.1 & 52.7 \\ 
        ~ & glove.6B.300d & \textbf{84.8} & 60.0 & \textbf{76.8} & \textbf{82.8} & 54.3 \\         
        ~ & bert-base-uncased & 79.0 (12) & 73.1 (10) & 67.3 (2) & 58.0 (9) & 61.5 (11) \\ 
        ~ & bert-large-uncased & 80.1 (16) & 74.8 (24) & 67.6 (1) & 64.7 (6) &  \textbf{67.2} (19) \\ 
        ~ & bert-base-multilingual-uncased & 84.4 (10) &  \textbf{76.0} (11) & 65.1 (0) & 71.5 (5) & 61.9 (2) \\ 
        ~ & roberta-base & 78.4 (11) & 75.2 (12) & 63.5 (0) & 69.9 (12) & 63.0 (12) \\ 
        ~ & roberta-large & 83.3 (19) & 75.3 (11) & 65.4 (2) &  \textbf{86.7} (23) & 60.2 (14) \\ 
        ~ & xlm-roberta-base & 76.2 (3) & 69.5 (4) & 58.9 (0) & 50.3 (10) & 59.7 (9) \\ 
        ~ & xlm-roberta-large & 79.6 (13) & 74.1 (13) & 64.6 (1) & 58.6 (4) & 56.2 (0) \\ 
        
        \hline
        \multirow{10}{1.5cm}{Max} & frequency & 80.7 & 46.4 & 57.2 & 42.5 & 47.9 \\ 
        ~ & fasttext.wiki & 82.0 & 54.3 & 74.9 & 47.7 & 56.0 \\ 
        ~ & glove.6B.300d & \textbf{89.4} & 58.0 & \textbf{76.0} & 63.4 & 55.8 \\         
        ~ & bert-base-uncased & 81.3 (12) & 66.3 (2) & 70.8 (2) & 62.3 (0) & 58.9 (12) \\ 
        ~ & bert-large-uncased & 81.8 (16) & 66.9 (6) & 68.4 (0) & 65.4 (3) &  \textbf{63.9} (14) \\ 
        ~ & bert-base-multilingual-uncased & 86.0 (7) & 65.3 (12) & 71.7 (0) & 69.5 (2) & 59.1 (12) \\ 
        ~ & roberta-base & 80.8 (11) & 66.6 (1) & 64.6 (0) & 56.6 (0) & 60.0 (10) \\ 
        ~ & roberta-large & 86.2 (19) &  \textbf{67.6} (4) & 66.2 (2) &  \textbf{71.7} (24) & 62.4 (23) \\ 
        ~ & xlm-roberta-base & 78.4 (2) & 59.9 (12) & 61.6 (0) & 68.2 (5) & 52.3 (7) \\ 
        ~ & xlm-roberta-large & 82.4 (13) & 65.3 (11) & 65.4 (1) & 52.9 (4) & 53.2 (10) \\ 
        \bottomrule
    \end{tabular}
    }
    \caption{Accuracy of different models under the \textbf{layer aggregation} setting. The optimal layer number for each contextualized LM is in brackets. The highest performance within each pooling method is in boldface.}
    \label{table:lm_perf_layeragg}
\end{table*}

Table~\ref{table:lm_perf_singlelayer} and Table~\ref{table:lm_perf_layeragg} show the detailed performance of specific LMs under the single-layer and the layer aggregation settings, respectively. From the results, we find that there is no consistent winner among all LMs. In terms of layers, on StylePPDB (formality short), the initial layers (0, 1, 2) are dominantly the best-performing ones across all settings. On the other datasets, there is no clear pattern in terms of which layers perform the best.

\subsection{Performance Across Layers Under Single-layer Setting}
\label{sec:appendix_layerperf_single}

\begin{figure*}
  \centering
  \begin{subfigure}[b]{0.3\textwidth}
    \includegraphics[width=\textwidth]{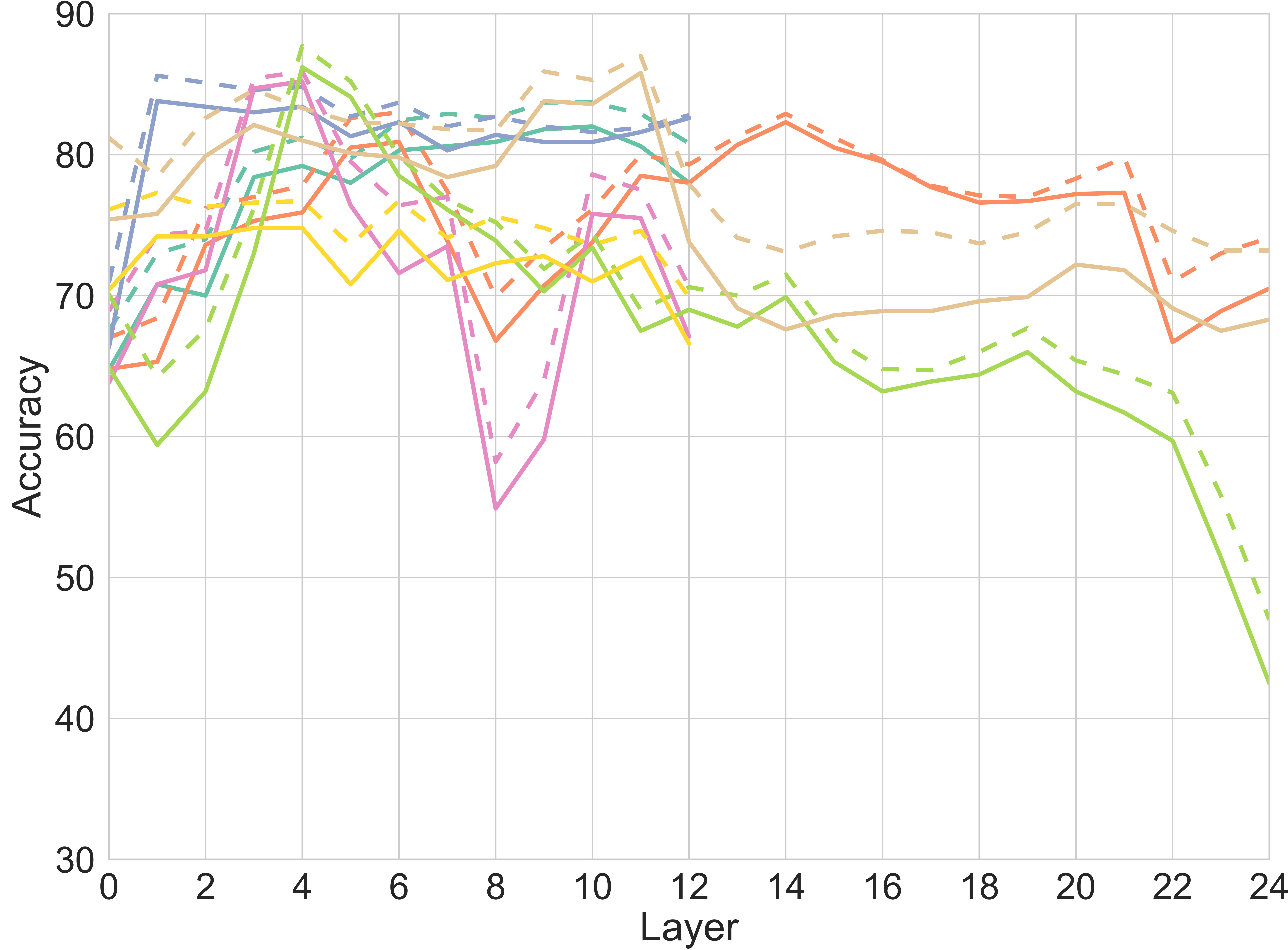}
    \caption{Complexity (short)}
  \end{subfigure}
  \hfill 
  \begin{subfigure}[b]{0.3\textwidth}
    \includegraphics[width=\textwidth]{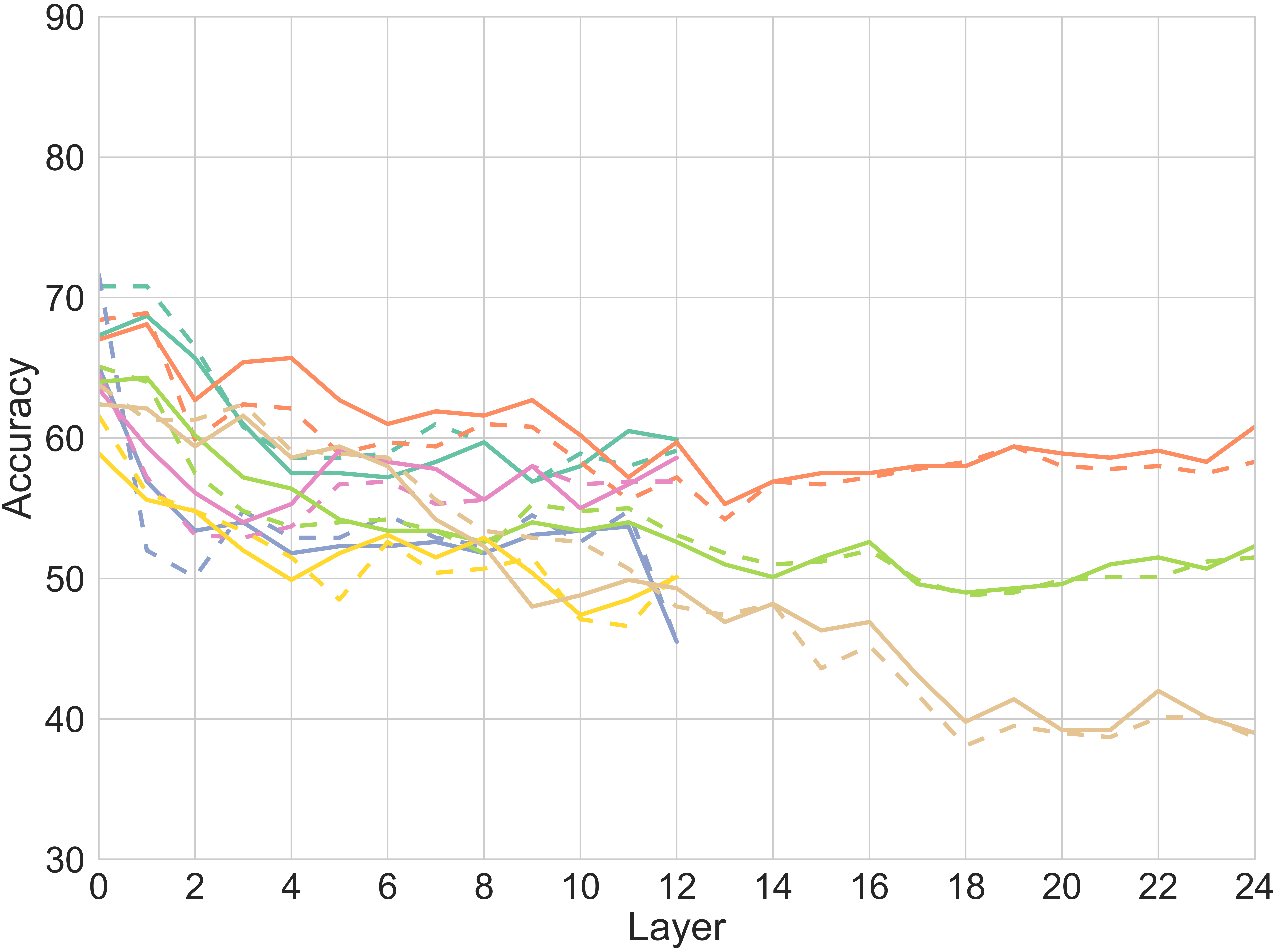}
    \caption{Formality (short)}
  \end{subfigure}
  \hfill 
  \begin{subfigure}[b]{0.3\textwidth}
    \raisebox{0.2cm}{\includegraphics[height=3.8cm]{img/layer_perf/legend.png}}
  \end{subfigure}

  \begin{subfigure}[b]{0.3\textwidth}
    \includegraphics[width=\textwidth]{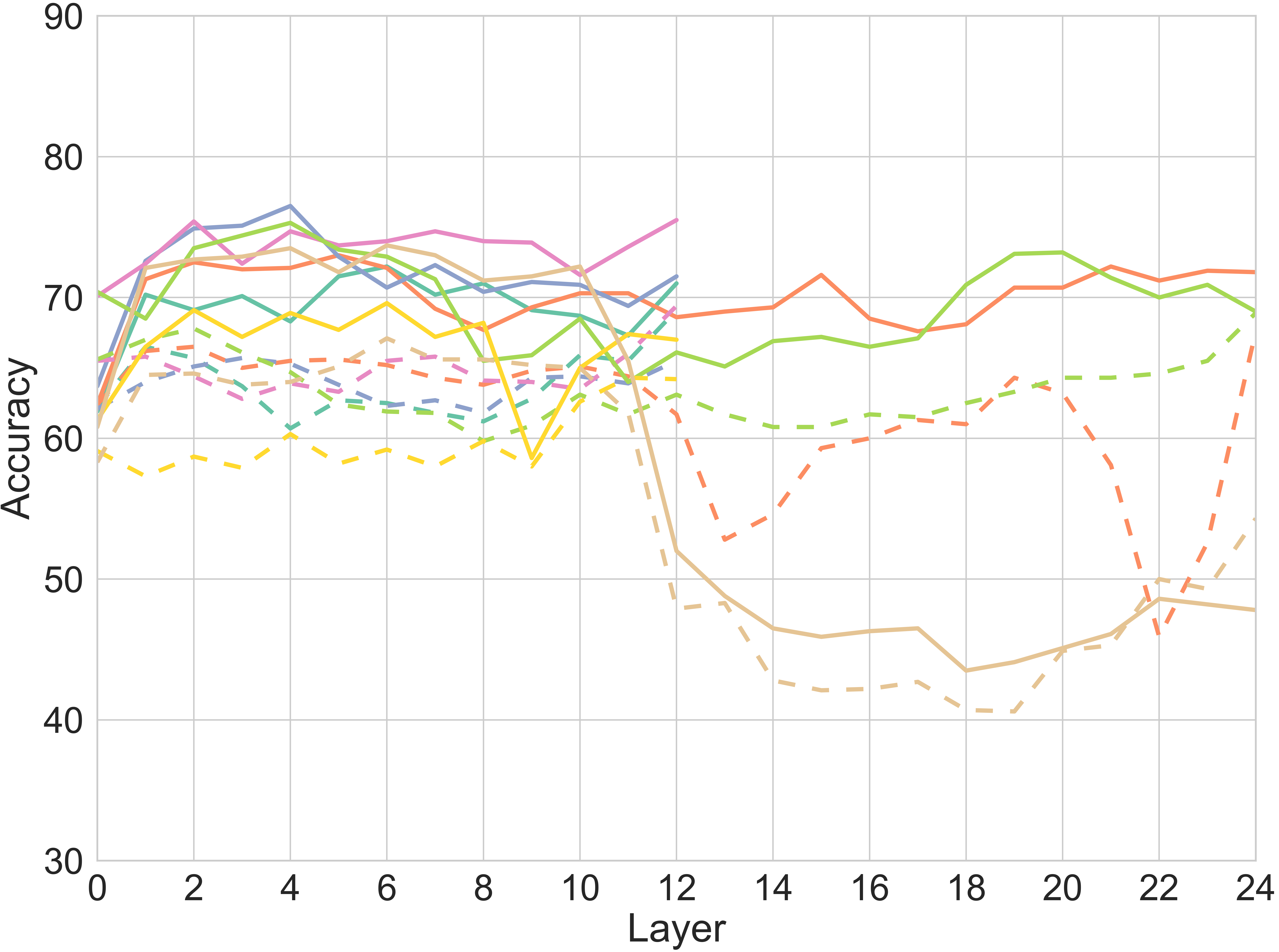}
    \caption{Complexity (long)}
  \end{subfigure}
  \hfill 
  \begin{subfigure}[b]{0.3\textwidth}
    \includegraphics[width=\textwidth]{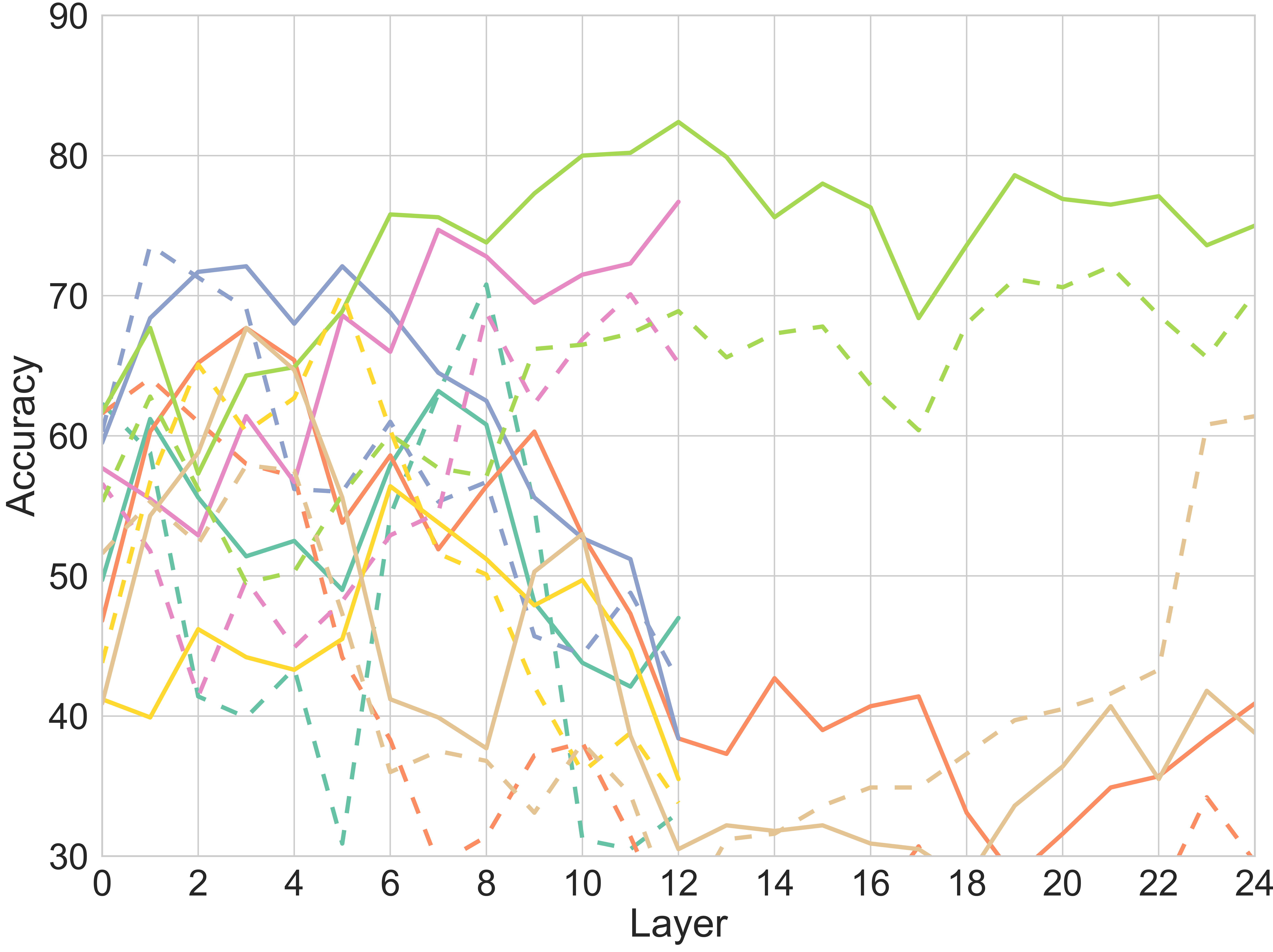}
    \caption{Formality (long)}
  \end{subfigure}
  \hfill   
  \begin{subfigure}[b]{0.3\textwidth}
    \includegraphics[width=\textwidth]{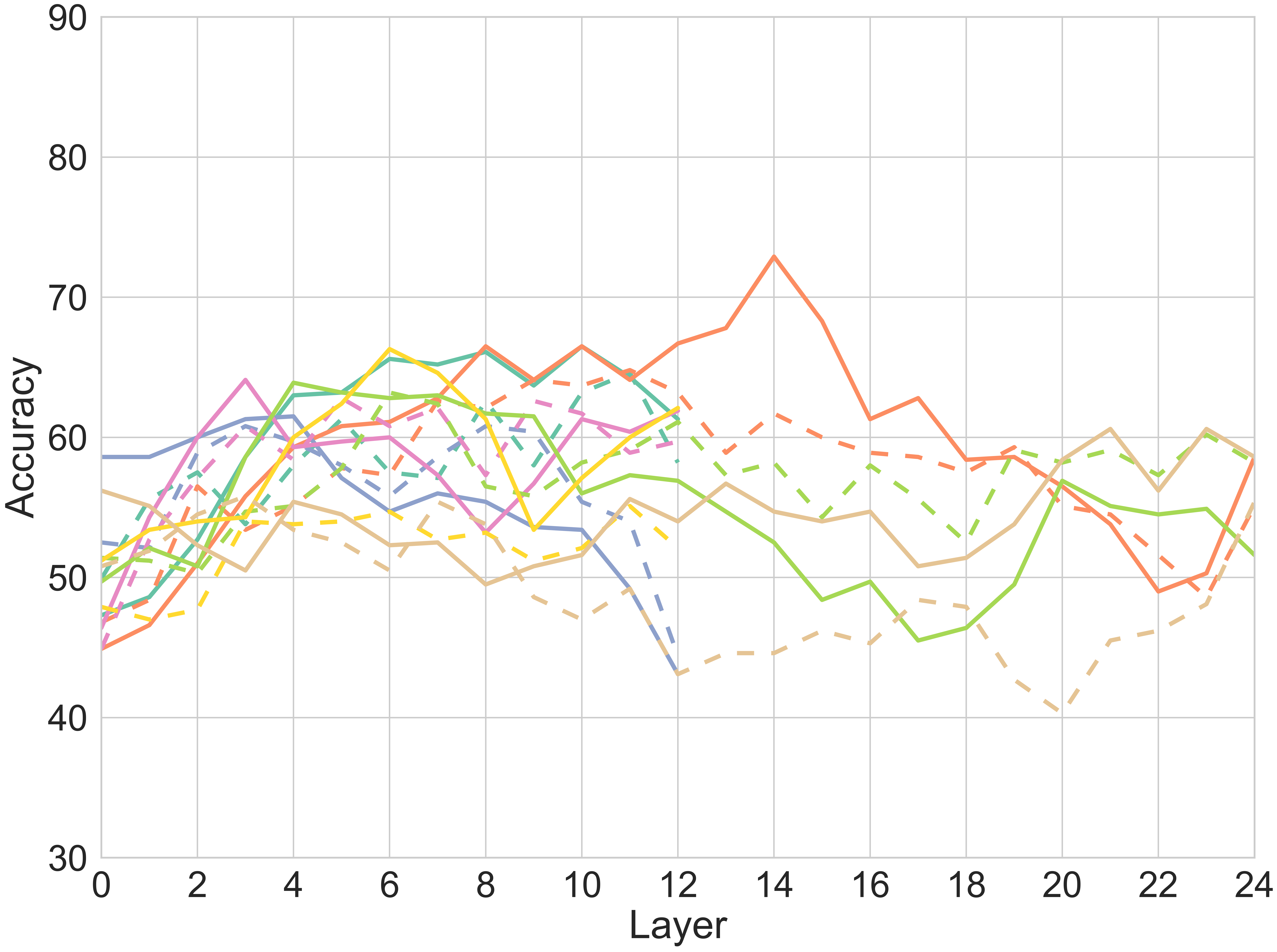}
    \caption{Figurativeness (long)}
  \end{subfigure}
  \hfill 
  \caption{Performance change across layers of different LMs (under the \textbf{single-layer} setting). }
  \label{fig:layer_perf_singlelayer}
\end{figure*}

Regarding the performance change across layers, in addition to the plots shown in Section~\ref{sec:perf_by_layer} under the layer aggregation setting, here we present the results under the single-layer setting in Figure~\ref{fig:layer_perf_singlelayer}. Compared to layer aggregation, the results here are noticeably more chaotic, exhibiting no clear general trends.

\subsection{Performance by Text Length Under Single-layer Setting}
\label{sec:appendix_lengthperf_single}

\begin{figure*}
  \centering
  
  \begin{subfigure}[b]{0.3\textwidth}
    \includegraphics[width=\textwidth]{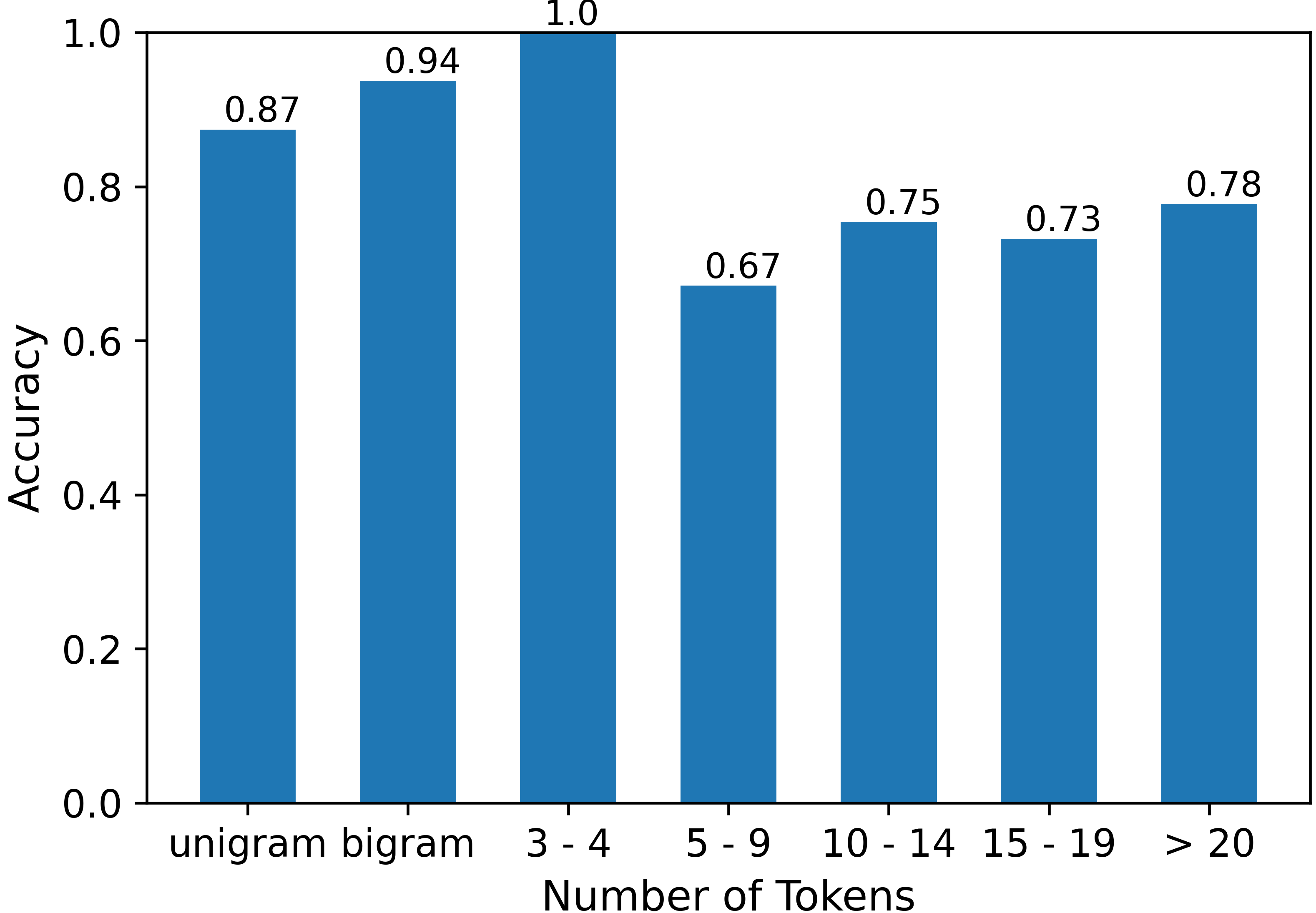}
    \caption{Complexity}
  \end{subfigure}
  \hfill 
  \begin{subfigure}[b]{0.3\textwidth}
    \includegraphics[width=\textwidth]{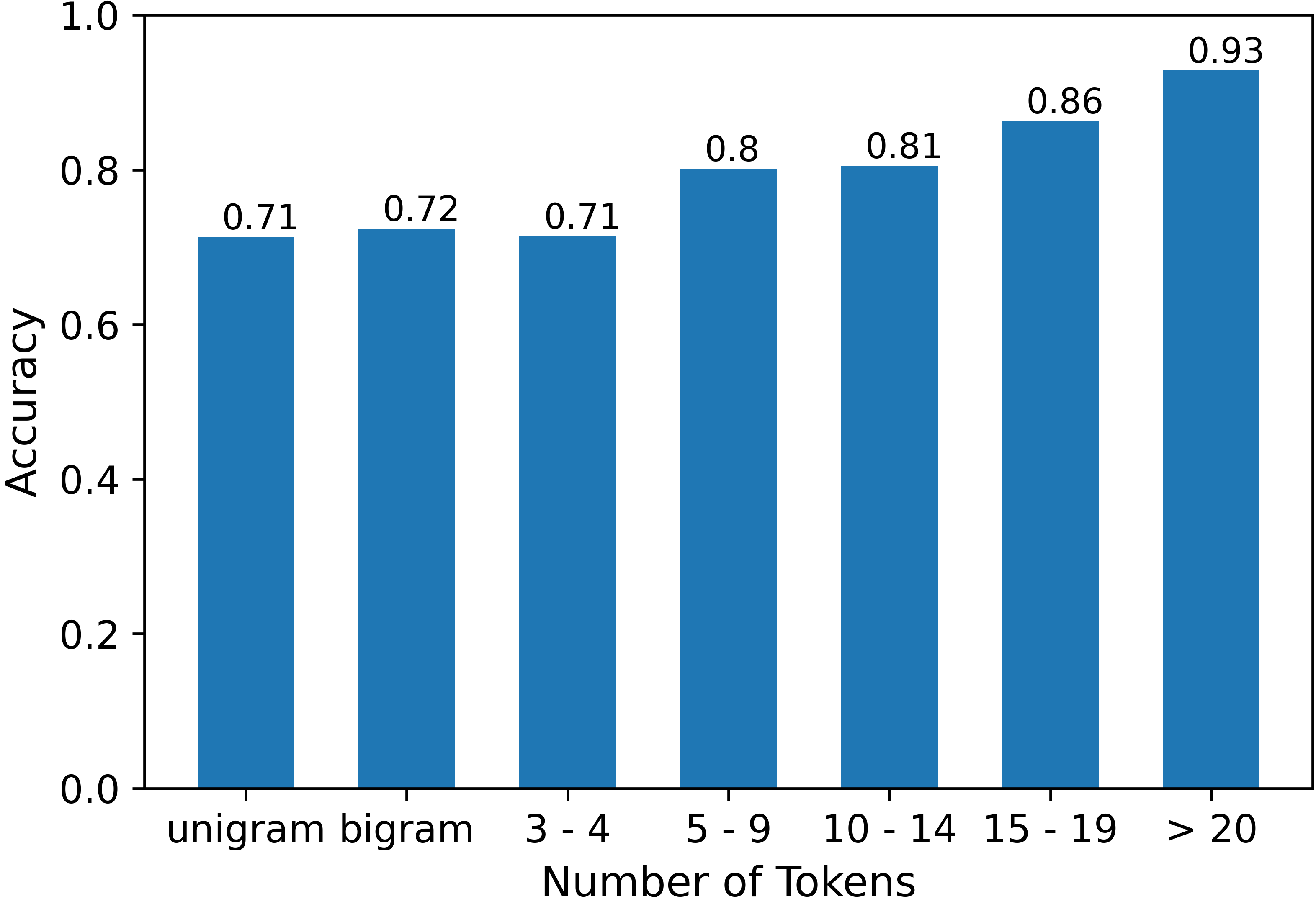}
    \caption{Formality}
  \end{subfigure}
  \hfill 
  \begin{subfigure}[b]{0.3\textwidth}
    \includegraphics[width=\textwidth]{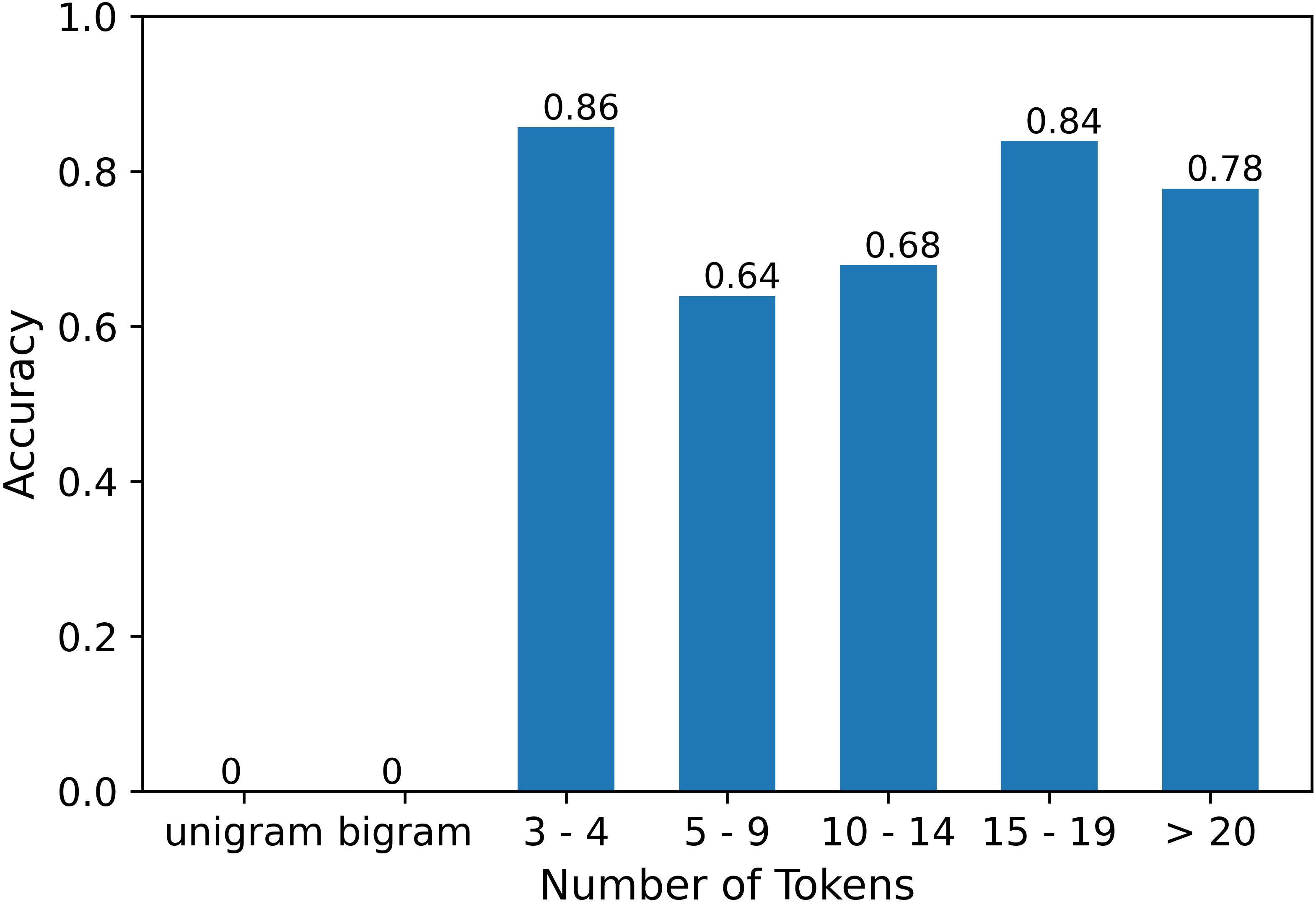}
    \caption{Figurativeness}
  \end{subfigure}
  
  \caption{Optimal performance over different bins of text length (under the \textbf{single-layer} setting).}
  \label{fig:acc_by_length_singlelayer}
\end{figure*}

Similarly, we also present the performance change by text length under the single-layer setting in Figure~\ref{fig:acc_by_length_singlelayer}, complementing the results under the layer aggregation setting in Section~\ref{sec:perf_by_length}. The trends are mostly similar between the two settings.

\subsection{Effect of Anisotropy Reduction}
\label{sec:appendix_anisotropy_comparison}

\begin{table*}
    \scalebox{0.65}{
    \begin{subtable}{\columnwidth}
        \centering
        \begin{tabular}{ll|cc|cc|c}
        \toprule
            \textbf{Pooling} & \textbf{Stats} & \multicolumn{2}{c|}{\textbf{Complexity}} & \multicolumn{2}{c|}{\textbf{Formality}} & \textbf{Figurativeness} \\
            ~ & ~ & short & long & short & long & long \\
             \hline
            \multirow{2}{1.4cm}{Mean} & 2 beats 1 (\%) & 25.2 & 21.3 & 44.9 & 40.2 & 74.0 \\ 
            ~ & acc gain & \cellcolor{pink!50} -5.9 & \cellcolor{pink!50} -5.0 & \cellcolor{pink!50} -0.2 & \cellcolor{pink!50} -8.2 & \cellcolor{green!20} 4.3 \\
            \hline
            \multirow{2}{1.4cm}{Max}  & 2 beats 1 (\%)& 18.9 & 28.3 & 48.0 & 44.1 & 59.1 \\
            ~ & acc gain & \cellcolor{pink!50} -7.4 & \cellcolor{pink!50} -2.2 & 0.0 & \cellcolor{pink!50} -0.3 & \cellcolor{green!20} 1.7 \\ 
            \hline
            \multirow{2}{1.4cm}{Average}  & 2 beats 1 (\%) & 22.0 & 24.8 & 46.5 & 42.1 & 66.5 \\ 
            ~ & acc gain & \cellcolor{pink!50} -6.7 & \cellcolor{pink!50} -3.6 & \cellcolor{pink!50} -0.1 & \cellcolor{pink!50} -4.2 & \cellcolor{green!20} 3.0 \\ 
            \bottomrule
        \end{tabular}
    \caption{All-but-the-top (single-layer)}
    \end{subtable}
    \hspace{4.5cm}
    \begin{subtable}{\columnwidth}
        \centering
        \begin{tabular}{ll|cc|cc|c}
        \toprule
            \textbf{Pooling} & \textbf{Stats} & \multicolumn{2}{c|}{\textbf{Complexity}} & \multicolumn{2}{c|}{\textbf{Formality}} & \textbf{Figurativeness} \\
            ~ & ~ & short & long & short & long & long \\
             \hline
           \multirow{2}{1.4cm}{Mean} & 2 beats 1 (\%) & 27.6 & 6.3 & 46.5 & 26.8 & 75.6 \\ 
            ~ & acc gain & \cellcolor{pink!50} -4.6 & \cellcolor{pink!50} -8.1 & \cellcolor{pink!50} -0.2 & \cellcolor{pink!50} -14.7 & \cellcolor{green!20} 5.6 \\ 
            \hline
            \multirow{2}{1.4cm}{Max}  & 2 beats 1 (\%)&  16.5 & 15.0 & 47.2 & 38.6 & 65.4 \\ 
            ~ & acc gain & \cellcolor{pink!50} -6.8 & \cellcolor{pink!50} -5.3 & \cellcolor{pink!50} -0.1 & \cellcolor{pink!50} -4.0 & \cellcolor{green!20} 3.1 \\ 
            \hline
            \multirow{2}{1.4cm}{Average}  & 2 beats 1 (\%) & 22.0 & 10.6 & 46.9 & 32.7 & 70.5 \\ 
            ~ & acc gain &  \cellcolor{pink!50} -5.7 & \cellcolor{pink!50} -6.7 & \cellcolor{pink!50} -0.1 & \cellcolor{pink!50} -9.4 & \cellcolor{green!20} 4.3 \\ 
            \bottomrule
        \end{tabular}
    \caption{All-but-the-top (layer aggregation)}
    \end{subtable}
    }

    \scalebox{0.65}{
    \begin{subtable}{\columnwidth}
        \centering
        \begin{tabular}{ll|cc|cc|c}
        \toprule
            \textbf{Pooling} & \textbf{Stats} & \multicolumn{2}{c|}{\textbf{Complexity}} & \multicolumn{2}{c|}{\textbf{Formality}} & \textbf{Figurativeness} \\
            ~ & ~ & short & long & short & long & long \\
             \hline
            \multirow{2}{1.4cm}{Mean} & 2 beats 1 (\%)  & 66.9 & 18.1 & 79.5 & 69.3 & 50.4 \\ 
            ~ & acc gain & \cellcolor{green!20} 3.9 & \cellcolor{pink!50} -3.6 & \cellcolor{green!20} 5.5 & \cellcolor{green!20} 9.3 & \cellcolor{pink!50} -0.7 \\ 
            \hline
            \multirow{2}{1.4cm}{Max}  & 2 beats 1 (\%)&  50.4 & 17.3 & 80.3 & 67.7 & 38.6 \\
            ~ & acc gain & \cellcolor{green!20} 1.6 & \cellcolor{pink!50} -4.8 & \cellcolor{green!20} 5.1 & \cellcolor{green!20} 12.1 & \cellcolor{pink!50} -2.1 \\ 
            \hline
            \multirow{2}{1.4cm}{Average}  & 2 beats 1 (\%) & 58.7 & 17.7 & 79.9 & 68.5 & 44.5 \\
            ~ & acc gain & \cellcolor{green!20}  2.7 & \cellcolor{pink!50} -4.2 & \cellcolor{green!20} 5.3 & \cellcolor{green!20} 10.7 & \cellcolor{pink!50} -1.4 \\ 
            \bottomrule
        \end{tabular}
    \caption{Standardization (single-layer)}        
    \end{subtable}
    \hspace{4.5cm}
    \begin{subtable}{\columnwidth}
        \centering
        \begin{tabular}{ll|cc|cc|c}
        \toprule
            \textbf{Pooling} & \textbf{Stats} & \multicolumn{2}{c|}{\textbf{Complexity}} & \multicolumn{2}{c|}{\textbf{Formality}} & \textbf{Figurativeness} \\
            ~ & ~ & short & long & short & long & long \\
             \hline
            \multirow{2}{1.4cm}{Mean} & 2 beats 1 (\%) & 86.6 & 15.0 & 93.7 & 60.6 & 72.4 \\ 
            ~ & acc gain & \cellcolor{green!20} 6.8 & \cellcolor{pink!50} -3.2 & \cellcolor{green!20} 7.2 & \cellcolor{green!20} 2.5 & \cellcolor{green!20} 2.5 \\ 
            \hline
            \multirow{2}{1.4cm}{Max}  & 2 beats 1 (\%)&  69.3 & 9.4 & 84.3 & 63.8 & 40.9 \\ 
            ~ & acc gain & \cellcolor{green!20} 4.0 & \cellcolor{pink!50} -6.1 & \cellcolor{green!20} 5.7 & \cellcolor{green!20} 7.6 & \cellcolor{pink!50} -0.8 \\ 
            \hline
            \multirow{2}{1.4cm}{Average}  & 2 beats 1 (\%) & 78.0 & 12.2 & 89.0 & 62.2 & 56.7 \\ 
            ~ & acc gain & \cellcolor{green!20} 5.4 & \cellcolor{pink!50} -4.7 & \cellcolor{green!20} 6.5 & \cellcolor{green!20} 5.0 & \cellcolor{green!20} 0.8 \\ 
            \bottomrule
        \end{tabular}
    \caption{Standardization (layer aggregation)}
    \end{subtable}
    }

    \scalebox{0.65}{
    \begin{subtable}{\columnwidth}
        \centering
        \begin{tabular}{ll|cc|cc|c}
        \toprule
            \textbf{Pooling} & \textbf{Stats} & \multicolumn{2}{c|}{\textbf{Complexity}} & \multicolumn{2}{c|}{\textbf{Formality}} & \textbf{Figurativeness} \\
            ~ & ~ & short & long & short & long & long \\
             \hline
            \multirow{2}{1.4cm}{Mean} & 2 beats 1 (\%) & 48.8 & 29.1 & 46.5 & 46.5 & 48.8 \\ 
            ~ & acc gain & \cellcolor{green!20} 0.1 & \cellcolor{pink!50} -1.1 & \cellcolor{pink!50} -0.4 & \cellcolor{green!20} 1.0 & \cellcolor{pink!50} -0.3 \\ 
            \hline
            \multirow{2}{1.4cm}{Max}  & 2 beats 1 (\%)&  47.2 & 38.6 & 46.5 & 63.8 & 52.0 \\ 
            ~ & acc gain & \cellcolor{pink!50} -0.3 & \cellcolor{pink!50} -0.8 & \cellcolor{pink!50} -0.3 & \cellcolor{green!20} 3.3 & \cellcolor{green!20} 0.4 \\ 
            \hline
            \multirow{2}{1.4cm}{Average}  & 2 beats 1 (\%) & 48.0 & 33.9 & 46.5 & 55.1 & 50.4 \\ 
            ~ & acc gain & \cellcolor{pink!50} -0.1 & \cellcolor{pink!50} -1.0 & \cellcolor{pink!50} -0.4 & \cellcolor{green!20} 2.1 & \cellcolor{green!20} 0.1 \\ 
            \bottomrule
        \end{tabular}
    \caption{Rank-based (single-layer)} 
    \end{subtable}
    \hspace{4.5cm}
    \begin{subtable}{\columnwidth}
        \centering
        \begin{tabular}{ll|cc|cc|c}
        \toprule
            \textbf{Pooling} & \textbf{Stats} & \multicolumn{2}{c|}{\textbf{Complexity}} & \multicolumn{2}{c|}{\textbf{Formality}} & \textbf{Figurativeness} \\
            ~ & ~ & short & long & short & long & long \\
             \hline
            \multirow{2}{1.4cm}{Mean} & 2 beats 1 (\%) &  54.3 & 32.3 & 33.1 & 36.2 & 54.3 \\ 

            ~ & acc gain & \cellcolor{pink!50} -0.2 & \cellcolor{pink!50} -1.8 & \cellcolor{pink!50} -0.9 & \cellcolor{green!20} 0.5 & \cellcolor{pink!50} -0.2 \\ 
            \hline
            \multirow{2}{1.4cm}{Max}  & 2 beats 1 (\%)&  54.3 & 48.0 & 43.3 & 71.7 & 62.2 \\ 
            ~ & acc gain & \cellcolor{pink!50} -0.4 & \cellcolor{green!20} 0.3 & \cellcolor{pink!50} -0.5 & \cellcolor{green!20} 3.2 & \cellcolor{green!20} 0.8 \\ 
            \hline
            \multirow{2}{1.4cm}{Average}  & 2 beats 1 (\%) & 54.3 & 40.2 & 38.2 & 53.9 & 58.3 \\ 
            ~ & acc gain &  \cellcolor{pink!50} -0.3 & \cellcolor{pink!50} -0.8 & \cellcolor{pink!50} -0.7 & \cellcolor{green!20} 1.9 & \cellcolor{green!20} 0.3 \\ 
            \bottomrule
        \end{tabular}
    \caption{Rank-based (layer aggregation)} 
    \end{subtable}
    }

    \caption{Effect of three different anisotropy reduction strategies: all-but-the-top, standardization, and rank-based (3 rows). Each strategy is evaluated under single-layer and layer aggregation settings (2 columns). In each table, ``2 beats 1 (\%)'' refers to the percentage of cases where the performance \textbf{with} the anisotropy reduction strategy is at least as high as the performance \textbf{without} it, under the same configuration (LM \& layer). ``Acc gain'' stands for the average accuracy gain of applying the anisotropy reduction strategy across all configurations. Positive accuracy gains are highlighted in green, negative ones in pink.}
    \label{table:anistropy_comparison}
\end{table*}

Table~\ref{table:anistropy_comparison} shows the effect of using the 3 different anisotropy reduction strategies, across all LM and layer configurations. All-but-the-top only works for figurativeness; rank-based only works for formality (long); and standardization works slightly more generally, for complexity (short), formality (short), and formality (long). Nevertheless, overall there is no strategy that works universally under every condition.

\end{document}